\documentclass[sn-mathphys,Numbered]{sn-jnl}

\usepackage{graphicx}%
\usepackage{multirow}%
\usepackage{amsmath,amssymb,amsfonts}%
\usepackage{amsthm}%
\usepackage{mathrsfs}%
\usepackage[title]{appendix}%
\usepackage{xcolor}%
\usepackage{textcomp}%
\usepackage{manyfoot}%
\usepackage{booktabs}%
\usepackage{algorithm}%
\usepackage{algorithmicx}%
\usepackage{algpseudocode}%
\usepackage{listings}%
\usepackage{times}
\usepackage{epsfig}
\usepackage{graphicx}
\usepackage{subcaption}
\usepackage{amsmath}
\usepackage{amssymb}
\usepackage{multirow}
\usepackage{diagbox}
\usepackage{colortbl}
\usepackage{booktabs}
\usepackage{pifont}
\usepackage{makecell}

\theoremstyle{thmstyleone}%
\theoremstyle{thmstyletwo}%

\theoremstyle{thmstylethree}%

\begin{document}

\title[Article Title]{Improved Region Proposal Network for Enhanced Few-Shot Object Detection}


\author[1]{\fnm{Zeyu} \sur{Shangguan}}\email{zshanggu@alumni.usc.edu}

\author[1]{\fnm{Mohammad} \sur{Rostami}}\email{rostamim@usc.edu}

\affil*[1]{\orgdiv{Department of Computer Science}, \orgname{University of Southern California}, \orgaddress{ \city{Los Angeles},  \state{CA}, \country{USA}}}


\abstract{Despite significant success of deep learning in object detection tasks, the standard training of deep neural networks requires access to a substantial quantity of annotated images across all classes. Data annotation is an arduous and time-consuming endeavor, particularly when dealing with infrequent objects.
Few-shot object detection (FSOD) methods have emerged as a solution to the limitations of classic object detection approaches based on deep learning. FSOD methods demonstrate remarkable performance by achieving robust object detection using a significantly smaller amount of training data. A challenge for FSOD is that instances from   novel classes that do not belong to the fixed set of training classes appear in the background and the base model may pick them up as potential objects. These objects behave similarly to label noise because they are classified as one of the training dataset classes, leading to FSOD performance degradation. We develop a semi-supervised algorithm to detect and then utilize these unlabeled novel objects as positive samples during the FSOD training stage to improve FSOD performance. Specifically, we develop a hierarchical ternary classification region proposal network (HTRPN) to localize the potential unlabeled novel objects and assign them new objectness labels to distinguish these objects from the base training dataset classes. Our improved hierarchical sampling strategy for the region proposal network (RPN) also boosts the perception ability of the object detection model for large objects. We test our approach and COCO and PASCAL VOC baselines that are commonly used in FSOD literature. Our experimental results indicate that our method is effective and outperforms the existing state-of-the-art (SOTA) FSOD methods. Our implementation is provided as a supplement to support reproducibility of the results ~\url{https://github.com/zshanggu/HTRPN}.\footnote{Early partial results of this work is presented in the  2023 ICCV Workshop on Visual Continual Learning~\cite{shangguan2023identification}.}}




\maketitle

\section{Introduction}

The integration of deep neural network (DNN) architectures into object detection has revolutionized the field, resulting in development of powerful models for accurately identifying and localizing objects of interest within an image. Object recognition DNNs  have significantly improved the ability to automatically determine the precise location and category of  objects of interest, providing invaluable insights for a wide range of applications. In the presence of abundant training data, object detection models that employ the region-based convolution neural networks (R-CNN) architecture~\cite{girshick2015fast,ren2015faster,he2017mask} have demonstrated remarkable perofrmnace in achieving high levels of accuracy across a broad spectrum of object detection tasks. While large-scale annotated training data can be a valuable resource for many applications, it can pose a significant challenge in certain domains, such as the analysis of miscellaneous diseases and industrial defect detection~\cite{zhou2017machine,rostami2018crowdsourcing}. Developing and curating high-quality annotated data requires extensive expertise in relevant fields, careful selection of relevant sources, and rigorous quality control measures to ensure accuracy and consistency across different datasets. Additionally, handling large volumes of data can require specialized hardware and software infrastructure, which may be costly and time-consuming to implement. In the presence of insufficient training data, DNNs   overfit and fail to generalize well during the testing page. Despite similarities between deep neural network and the nervous sytem\cite{morgenstern2014properties,schyns2022degrees}, humans possess an exceptional ability to quickly and accurately classify novel object classes based on a handful of samples~\cite{xian2018zero,wang2020generalizing,rostami2022zero}. This ability has motivated the development of models that can learn object classes with only a few samples, known as few-shot object detection (FSOD)~\cite{Han22a,Kohler22,rostami2019deep,Huang22,Sun21,Zhang22,Wang20,Kaul22}. FSOD methods have become increasingly popular due to their potential to generalize well on unseen data, their efficiency in handling limited training examples, and boradening the applicability of deep learning.

\begin{figure}[t]
    \centering
    \includegraphics[width=130mm]{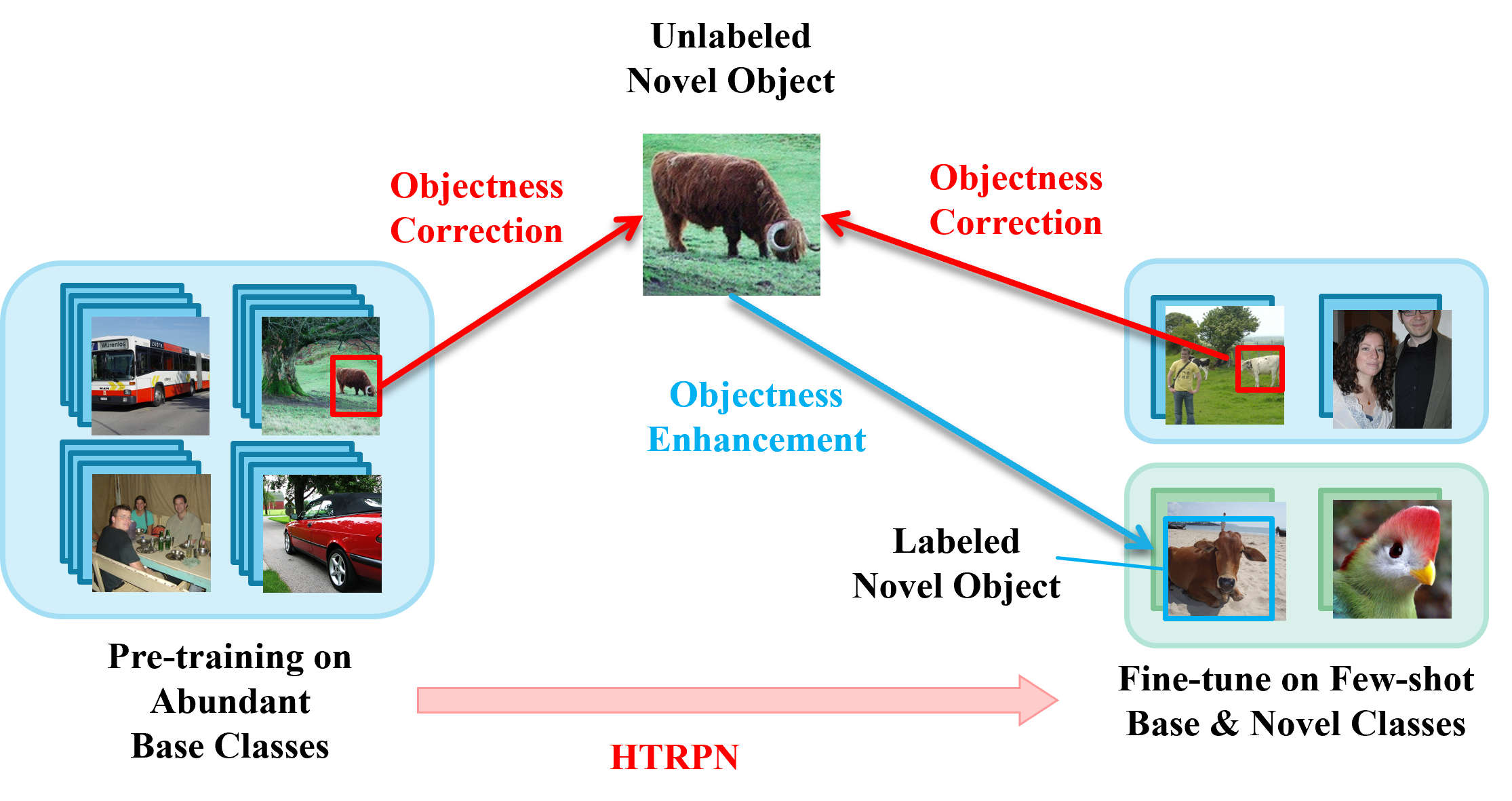}
    \caption{\small The dominant FSOD pipline: FSOD methods pre-train a model on abundant base classes and then fine-tune it on both the base and the novel few-shot classes. Some instances that remain unlabeled are typically ignored during training, resulting in a significant impact on the convergence of   model training. To address this issue, our proposed HTRPN architecture takes into account the importance of these unlabeled instances and incorporates them into the training process. By doing so, we can improve the overall performance of the model and ensure its ability to generalize well to unseen data.}
    \label{fig:intro}
\end{figure}

Currently, the predominant approach to FSOD involves pre-training a suitable model using a set of base classes that possess ample training data. This process enables the model to acquire a robust understanding of the basic helpful features and patterns within these classes. Once trained, the model is then fine-tuned on both the base classes and new classes that are more challenging to learn due to their limited sample size. The fine-tuning process allows the model to adapt to new classes and improve its performance on previously unseen classes during pre-training (see Figure~\ref{fig:intro}).     
In other words, it is assumed that classes of interest can be divided into abundant base classes and few-shot   classes. 
The fundamental methodology employed in FSOD is to benefit from ideas in transfer learning or meta-learning to learn novel classes~\cite{romera2015embarrassingly,kolouri2018joint,soh2020meta}, while preserving satisfactory performance levels in the base categories. This approach entails utilizing the knowledge acquired during the pre-training phase to learn new classes, while simultaneously ensuring that the model's recongition ability in its original base classes remains intact. Despite advances in FSOD, most existing   state-of-the-art (SOTA) techniques   fail to yield favorable outcomes when applied to few-shot classes that share similarities with the base classes. For example, we can see in Figure~\ref{fig:intro}  that some novel classes (e.g., the cow in the input image) might encounter objectness inconsistency, i.e., some of the instances remain unlabeled in the base class images and are therefore treated as background objects, while the labeled instances are treated as foreground objects.  Potential reasons for this performance gap include the confusion between visually similar categories, incorrect annotations (label noise), the existence of unseen novel objects during training, etc. Recent FSOD  methods have focused on addressing these challenges to improve FSOD performance. 

In this paper, we focus on a challenge that has not been recognized and  addressed in the FODL literature.
Specifically, we study the phenomenon that unlabeled novel object classes that do not belong to either of the base or the few-shot   classes can appear in the training data. For example, we see in Figure.~\ref{fig:intro} that among base-class training samples, there are a number of objects that remain unlabeled, such as the cow in the image. These unlabeled objects can potentially belong to unseen novel classes. Our experiments demonstrate that this phenomenon exists in both PASCAL VOC~\cite{Everingham10} and COCO~\cite{Lin14} datasets. Similarly, this situation can also ocurr in real-word industrial applications. 
This phenomenon leads to the objectness inconsistency for the model when recognizing the novel objects: for the novel class, objects are treated as background if their annotations are missing, but they are treated as foreground when they are labeled. In other word, these instances lead to inconsistencies within each concept class and degrade the performance similar to the effect of concept shift~\cite{adler2020cross,rostami2021cognitively,rostami2021lifelong}. Such nonconformity of foreground and background confuses the model when predicting the objectness for these instances and make the model hard to converge and degrades detection accuracy. Additionally, when these instances are identified as one of the training set classes, they serve as noisy labels.

To tackle the above challenge, we develop a semi-supervised learning method to utilize the potential novel objects that appear during training to improve the ability of the model to recognize novel classes to avoid confusing them with instances of training classes. To this end, we first demonstrate the possibility of detecting these unlabeled objects, despite the fact that they are novel, i.e., labeled instances of these instances do not exist in the training dataset. Our experiment indicates that some unlabeled class objects are likely to be recognized if they are similar to the training base and novel classes. We collect the unlabeled novel objects from the background proposals by determining whether they are predicted as known classes, and then we give these proposals an extra objectness label in the region proposal network (RPN) so that the model can learn to identify them and distinguish them from the base classes. We also analyze the defect of the standard RPN in detecting objects of different sizes during training and propose a more balanced RPN sampling method so that objects are treated equally in all scales. We provide extensive experimental results to demonstrate the effectiveness of our method on the PASCAL VOC and COCO datasets.
Our contributions include:

\begin{itemize}
    \item In order to improve the performance of anchor generation, we modify the anchor sampling strategy by selecting anchors uniformly from various layers of the feature pyramid layer in the R-CNN architecture. We note that objects with relatively larger sizes have higher possibilities to be observed.
    
    \item We design a ternary objectness classification in the RPN layer which enables the model to recognize potential novel class objects to improve consistency. 
    
    \item We use contrastive learning in the RPN layer to distinguish between the positive and the negative anchors to mitigate the negative effects of novel objects on FSOD perofrmnace.
\end{itemize}

\section{Related works}

In our formulation, we assume that there are three classes in an FSOD setting: base classes, seen novel classes, and unseen novel classes. The base and the seen novel classes form the training dataset together and the goal is to train a model to indetify all of them. For base classes, we have sufficient instances of annotated data but for seen novel classes, we have a few samples per class. Most works in FSOD only consider these classes for evaluation. The unseen novel classes are not included in the training data but emerge as novel classes in the background due to missing explicit label. Instances of these classes exist in background without being included in the annotated samples.

\textbf{Object detection}
In the realm of object detection, there exist two primary paradigms to design suitable architectures: one-stage and two-stage networks. Two-stage networks, such as R-CNN~\cite{Girshick14}, R-FCN~\cite{Girshick14}, Fast-RCNN~\cite{girshick2015fast}, and Faster-RCNN~\cite{ren2015faster}, are commonly employed for their simplicity and efficiency. Two-stage object detection networks offer more flexibility and accuracy but come with increased training and inference complexity. In these networks, a region proposal network (RPN) is used to generate a set of potential object regions, which are then passed through a second stage classifier to produce instance-level class probabilities and a box regression head to give the precise location for each object. This approach allows for better handling of objects with varying sizes, shapes, and occlusions, as well as the ability to incorporate context information from neighboring regions. One-stage object detection networks, such as SSD ~\cite{Liu16}, YOLO series ~\cite{Redmon16, Redmon18}, and Overfeat ~\cite{Sermanet14}, have gained significant attention due to their efficiency and simplicity. These networks estimate both the category and the location of an object directly from the backbone network without the need for a RPN. By bypassing the RPN stage, one-stage networks can achieve faster inference times and lower resource requirements~\cite{Kohler22}.In the field of FSOD, we prioritize the accuracy of object detection as a critical issue. To assess the performance of our FSOD models, we have chosen to use the widely-recognized two-stage approach, which has been extensively studied and validated in previous works.

\textbf{Few-shot object detection} (FSOD), is a case of object detection and a subfield of few-shot learning ~\cite{ravi2016optimization,snell2017prototypical,sung2018learning,rostami2019sar,wang2020generalizing,} that specifically deals with the task of detecting objects in new scenes or conditions where only a handful of annotated examples are available for training. In general, few-shot object detection involves two key challenges: (1) the scarcity of training data, which can limit the accuracy and robustness of the model and (2) the need to generalize to novel objects, which requires the model to be able to recognize and classify objects that it has not seen before. To address these challenges, researchers have proposed a range of techniques. Two such approaches that have gained significant attention are meta-learning~\cite{sun2019meta,chen2021meta,goldblum2020adversarially} and transfer learning~\cite{mirtaheri2020one,gupta2020effective,lai2021graph}. Meta-learning   focuses on learning how to learn and specifically is inspired by studying how humans acquire knowledge and adapting this approach to the specific task of few-shot object detection. This approach involves developing models that can learn to learn from a small number of scenarios, allowing them to adapt quickly to new situations and improve their accuracy over time. Transfer learning, on the other hand, involves leveraging knowledge from related tasks with more data to improve the performance of novel object detection, even though there are only a few examples available for training and potentially no samples~\cite{rostami2020using}. Recent research on few-shot learning has demonstrated that sufficient pre-training is non-negligible and could significantly improve the recognition ability of the model when transferring to a new few-shot task, especially with huge models as backbone such as vision transformers~\cite{Hu22}.

The typical approach to address FSOD involves two phases: pre-training and fine-tuning. During the pre-training phase, the model is   trained on the set of base classes for which we possess a sufficient number of annotated data points. This stage allows the model to learn extracting generalizable features and patterns that can be applied to identify novel classes. Once the pre-training phase is complete, the model is then fine-tuned on a subset of novel classes, each with a limited number of samples~\cite{Wang20}. The purpose of the fine-tuning step is to further adapt the model to the specific characteristics of the unseen novel classes, ensuring that it can accurately classify them using the learned features and patterns from the pre-training phase.  Meta-learning and transfer learning, mentioned above, are two major end-to-end approaches that can be helpful in the fine-tuning stage,. In the context of knowledge representation and inference, meta-learning can be used to build an inquiry set and a support set for a $k$-way $n$-shot setting. A $k$-way $n$-shot setting involves creating a support set with $k$ categories and $n$ samples in each category. The goal is to train a model that can accurately classify an inquiry instance into its corresponding category based on the support set. Meta-learning based method such as Attention-RPN~\cite{Fan20} add an attention module on RPN and a detector to match relationships between query and support proposal pairs. Meta Faster R-CNN~\cite{Han22a} proposes a two-stage coarse-to-fine prototype matching network to optimize the region proposals by fusing the features of support and query instances. FCT~\cite{Han22b} introduces pyramid vision transformer as backbone, and applies a cross-attention head to aggregate the K-V matrices upon the query and support features, which can effectively speed up the training procedure and bridge the gap between the query and support branch. Sylph~\cite{Yin22} proposes a few-shot hypernetwork that contains a code generator that could better presents the average feature of the support instances. A hypernetwork are normally small networks ~\cite{von2020continual,jin2021learn,chandra2023continual} using the notion of adapters~\cite{srinivasan2023i2i,li2022cross}.

The typical approach to address FSOD involves two phases: pre-training and fine-tuning. During the pre-training phase, the model is   trained on the set of base classes for which we possess a sufficient number of annotated data points. This stage allows the model to learn extracting generalizable features and patterns that can be applied to identify novel classes. Once the pre-training phase is complete, the model is then fine-tuned on a subset of novel classes, each with a limited number of samples~\cite{Wang20}. The purpose of the fine-tuning step is to further adapt the model to the specific characteristics of the unseen novel classes, ensuring that it can accurately classify them using the learned features and patterns from the pre-training phase.  Meta-learning and transfer learning, mentioned above, are two major end-to-end approaches that can be helpful in the fine-tuning stage,. In the context of knowledge representation and inference, meta-learning can be used to build an inquiry set and a support set for a $k$-way $n$-shot setting. A $k$-way $n$-shot setting involves creating a support set with $k$ categories and $n$ samples in each category. The goal is to train a model that can accurately classify an inquiry instance into its corresponding category based on the support set. Meta-learning based method such as Attention-RPN~\cite{Fan20} add an attention module on RPN and a detector to match relationships between query and support proposal pairs. Meta Faster R-CNN~\cite{Han22a} proposes a two-stage coarse-to-fine prototype matching network to optimize the region proposals by fusing the features of support and query instances. FCT~\cite{Han22b} introduces pyramid vision transformer as backbone, and applies a cross-attention head to aggregate the K-V matrices upon the query and support features, which can effectively speed up the training procedure and bridge the gap between the query and support branch. Sylph~\cite{Yin22} proposes a few-shot hypernetwork that contains a code generator that could better presents the average feature of the support instances. A hypernetwork are normally small networks ~\cite{von2020continual,jin2021learn,chandra2023continual} using the notion of adapters~\cite{zhang2022tip,li2022cross,srinivasan2023i2i}.

In contrast to meta-learning methods, transfer learning-based techniques   utilize  pre-trained weights that have been trained on a large corpus of data. These pre-trained weights are then fine-tuned on the novel seen classes, where the model is retrained on a smaller dataset. TFA~\cite{Wang20} implements a standard fine-tune based FSOD benchmark based on two-stage Faster R-CNN, and demonstrate that fine-tune based method has comparable performance comparing to meta-learning based method and have simpler structure. FSCE~\cite{Sun21} further improves TFA by adding a contrastive learning module to achieve more balanced classification distance when fine-tuning on novel data. Retentive R-CNN~\cite{Fan21} demonstrates that RPN is not ideally class-agnostic and therefore presents debiased RPN to eliminate the effect from the pre-trained RPN. 
Adaptive R-CNN \cite{wang2019few}  mitigate the effect of  domain shift~\cite{rostami2023overcoming} on perofrmnace degradation using a pairing mechanism   to alleviate the issue of insufficient   samples. DeFRCN~\cite{qiao21} decouples the classification and box regression head during fine-tuning through a gradient decoupled layer since the former is translation-invariant and the latter is translation-covariant, and demonstrates its effectiveness. MFDC~\cite{Wu22} suggests that pre-trained model are not completely class-agnostic and not suitable to be used directly while fine-tuning. Therefore, they implemented a distillation framework so that the network could remember the key information only from the data-sufficient pre-training process. We select the fine-tuning based approach as our baseline in this paper because it is simple yet effective, and consumes less computational resources, given the limitations we have.

\textbf{Unseen novel objects}
In an object detection problem, the set of the base and the seen novel classes are assumed to be a closed set. However, there may be potential novel unseen objects in the training dataset that do not belong to the initial set of classes. These objects naturally are classified as one of seen classes and hence, there has been an interest to mitigate the adverse effects of these objects~\cite{rostami2021detection}. Semi-supervised object detection network is a potential solution for this problem which utilizes the challenging samples ~\cite{Rosenberg05, Liu21, Xu21}. Kaul et al. ~\cite{Kaul22} build a class-specific self-supervised label verification model to identify candidates of unlabeled (unseen) objects and give them pseudo-annotations. The model is then retrained with these pseudo-annotated samples to improve the object-detecting accuracy. However, this method requires two rounds of training and requires extra effort to adapt to other categories. Li et al. ~\cite{Li21c} propose a distractor utilization loss by giving the distractor proposals a pseudo-label during fine-tuning. This method is used only in the fine-tuning stage and hence, the objectness inconsistency from the pre-training stage is not addressed. Motivated by these drawbacks, we propose employing the unlabeled potential objects that belong to the unseen classes to ameliorate the detrimental impact of seen novel objects.

\textbf{Contrastive learning}
 can be used to enlarge the inter-class distances and narrow down the intra-class distances for classification tasks to enhance data representations~\cite{gao2021contrastive,liu2021learning,jian2023unsupervised}. Contrastive learning has been applied to many classification tasks in topics such as visual recognition ~\cite{Luo21, Wang21a}, semantic segmentation ~\cite{Wang21b}, super-resolution ~\cite{Wang23}, and natural language processing ~\cite{Fang05, Chi20}. Supervised contrastive learning is a classic form of contrastive learning that involves the use of labeled samples. In this approach, an encoder is employed to extract the distinctive features for each sample, and then a contrastive loss is constructed by assessing the consistency between the labels of sample pairs and their distance metric. Specifically, sample pairs with different labels but smaller feature distance are expected to have higher loss values, while those with similar labels and larger feature distances are expected to have lower loss values. Supervised contrastive learning in few-shot object detection is introduced by FSCE~\cite{Sun21} to better distinguish similar categories at   instance level during fine-tuning, a contrastive learning module is added parallel to the box regression layer and classification layer so that the inter/intra-class distances of the predicted instances could be well-balanced. We benefit from this   strategy in our work by applying contrastive learning to the region proposal (RPN) layer. The region proposals   suffer from severe foreground-background unbalancing, therefore,   contrastive learning would be a effective strategy in here to improve the objectness evaluation ability of the RPN layer.

\section{Problem Description}
\label{sec: Problem Description}

We formulate the problem of FSOD following a standard setting in the literature~\cite{Kang19, Wang20}. We use the Faster R-CNN network as the object detection model and adhere to the same evaluation methodology for FSOD established by Wang et al.~\cite{Wang20}. According to this formulation, the training dataset classes is split into two categories: the base and the novel seen classes. The base classes ($C_B$) are the ones that we have a sufficient number of images and instances for in our dataset, which means that they are well-represented and can be used as a basis for training the model to extract descriptive features. On the other hand, the novel seen classes ($C_N$) are infrequent classes for which we only have a few training samples in the dataset. As a result, the classes are less representative in the training dataset are and harder to analyze. Additionally, we consider this splitting to be exclusive, i.e., $C_B\cap C_N = \varnothing$. An $n$-shot learning scenario refers to the situation where we have access to $n$ instances per each seen novel category. During the pre-training stage, the model is trained exclusively on the base classes $C_B$ and then is evaluated solely on their corresponding test split set. This stage allows the model to learn the generalizable features and patterns of the base classes without being distracted by the novel classes. In the second stage, we focus on learning the seen novel classes. The goal is to use the learned feature extraction ability to learn the novel seen classes to avoid learning them from scratch.  During this stage, we fine-tune the pre-trained model on a smaller dataset containing the novel classes that were not present during the pre-training stage. A side effect can be performance degradation on the base classes because of the model updates. This process is called catastrophic forgetting. To prevent catastrophic forgetting~\cite{kirkpatrick2017overcoming,rostami2019complementary,rostami2020generative,shi2021overcoming,yap2021addressing} in the pre-trained model and ensure that it retains the knowledge learned about the base classes to continue performing well on these classes, we fine-tune the model on both the seen novel classes and the base classes $C_N\cup C_B$. Once the fine-tuning is complete, we test the model on both sets of classes to evaluate its performance and ensure that it can accurately detect and localize objects in both the base and novel classes.

The R-CNN architecture has been widely used for object detection tasks due to its effectiveness in identifying objects from images. In our work, we propose an improvement to the base R-CNN architecture that allows us to identify novel unseen classes as instances that do not belong to the seen classes. The R-CNN architecture is designed to identify objects in images by first extracting five scaled feature maps ($p2\sim p6$) from the input image using its feature pyramid network (FPN). These feature maps are then passed through the region proposal network (RPN) where size-fixed anchors are applied to predict the objectness scores and the coarse bounding box for each proposal region to rule whether the region contains an object. The objectness score ($obj\{obj_{pre}, obj_{gt}, iou_{gt}^a\}$) represents the likelihood of an object being present in a given region and ranges from 0 to 1, where $obj_{pre}$ is the predicted objectness score. the ground truth value $obj_{gt}=0$ indicates a non-object and $obj_{gt}=1$ represents a true object, and $iou_{gt}^a$ represents the intersection over the union of an anchor with its ground truth box. The coarse bounding boxes (i.e., $bbox_c$) represent the most likely proposal boxes for objects based on the proposals made by the RPN. The proposal boxes ($Prop\{obj, bbox_c, iou_{gt}^p\}$, where $iou_{gt}^p$ represents the intersection over the union of a proposal box with its ground truth box) is the input of the region of interest pooling layer (RoI pooling).
Anchors with an IoU greater than 0.7 ($iou_{gt}^a > 0.7$) are considered active anchors ($A_a$) and their corresponding proposals are called positive proposals ($Prop_p$). On the other hand, anchors with an IoU less than 0.3 ($iou_{gt}^a < 0.3$) are considered negative anchors ($A_n$) and their corresponding proposals are called negative proposals ($Prop_n$). Other proposals are ignored and would not be considered in the final object detection results.

After generating the objectness scores and bounding boxes using the RPN, the network proceed to use the RoI pooling layer to further refine the predictions and come up with the class label for the detected object. The RoI pooling layer takes the proposal boxes predicted by the RPN and performs pooling on their corresponding feature maps. These features are then used to predict the instance-level classification (i.e., $cls^i$, where $i$ is the classification index) and refined bounding box (i.e., $bbox_r$) for each object in the image. This instance-level classification is used to determine the class of the object being detected, while the refined bounding box provides more accurate information about the location and size of the object in the image. By combining these two predictions, we can effectively detect objects ($Obj\{cls^i, bbox_r\}$)  and their locations in the input images with high accuracy.

\begin{figure}[t]
	\centering
	\begin{subfigure}{0.45\linewidth}
		\centering
		\includegraphics[width=0.9\linewidth]{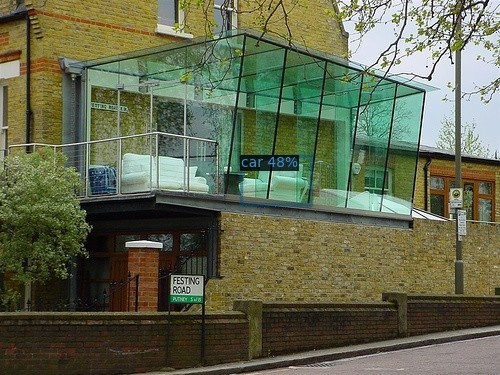}
		\caption{}
		\label{figure2_a}
	\end{subfigure}
	\centering
	\begin{subfigure}{0.45\linewidth}
		\centering
		\includegraphics[width=0.9\linewidth]{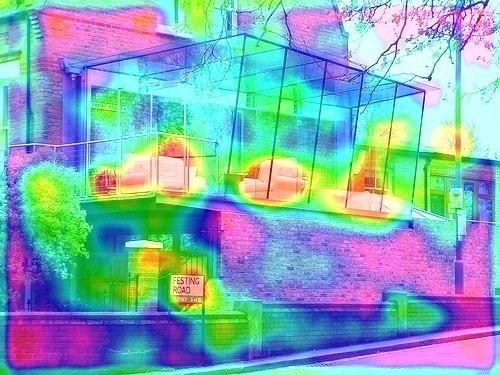}
		\caption{}
		\label{figure2_b}
	\end{subfigure}
	\centering
	\begin{subfigure}{0.45\linewidth}
		\centering
		\includegraphics[width=0.9\linewidth]{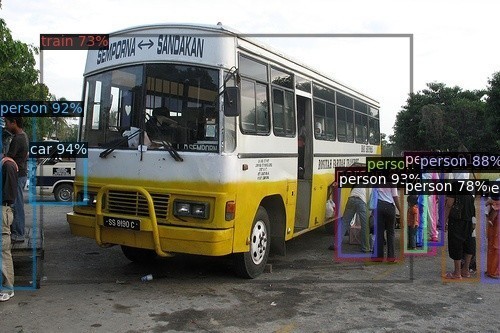}
		\caption{}
		\label{figure2_c}
	\end{subfigure}
	\centering
	\begin{subfigure}{0.45\linewidth}
		\centering
		\includegraphics[width=0.9\linewidth]{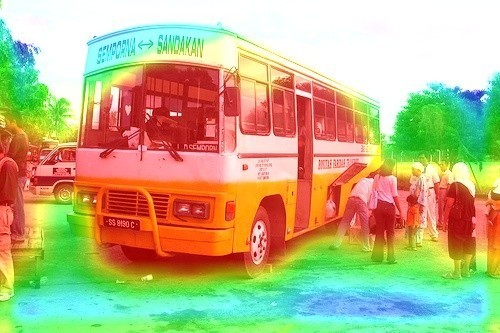}
		\caption{}
		\label{figure2_d}
	\end{subfigure}
	\caption{Grad-CAM visualization of R-CNN attention mao for a sample image: The left images in each row indicate the predicted boxes with the class names and confidence, and the right images are the feature maps of the RPN layer. The color of Grad-CAM represents the importance of each pixel in the heatmap, with red indicating the highest importance and blue indicating the lowest importance.This visualization helps us to understand the predictions made by the model and identify areas of interest in images. Upon close inspection, we can see misclassified instances in these examples that are classified as instances of training set classes. Our goal is to enable the model to distinguish instances of novel unseen objects from the seen objects in the training dataset to improve the quality of FSOD.}
	\label{fig:visualization}
\end{figure}

One of the challenges in object detection is that novel instances (${C_N}^{pn}$), which are not present in the training dataset and have not been seen before, can appear in the background or other non-overlapping regions of an image (see Figure~\ref{fig:visualization}). The reason is that there are many potential classes in the dataset that we have not included in either the base classes or the unseen novel classes. Instances of these classes can potentially be  detected in the first stage of object detection because these objects may be treated as $Prop_n$ and with its ground truth objectness ${obj_{gt}}^{pn}=0$. On the contrary, they would be treated as $Prop_p$ with ground truth objectness ${obj_{gt}}^{pn}=1$ if it is labeled as such by the model. These instances can significantly confuse the model when adapting the model for learning the novel unseen classes because they will be labeled as a class present in the training set. We argue that if the unlabeled potential novel object can be distinguished from the $Prop_n$, then its objectness could be modified as a foreground object. This process would eliminate the inconsistency of the objectness score and improve the performance of the model.
In other words, we propose to reduce an effect similar to noisy labels as these objects would be objects with wrong labels, leading to confusion in the model and performance degradation.

\section{Proposed solution}
\label{sec: Proposed Solution}
To propose our solution, we first demonstrate that there are instances of novel unseen classes that can be encountered in the training stage such that they are ruled as objects and then labeled as instances of seen classes. We then investigate the relationship between the number of anchors and the size of objects in each feature layer and provide a more effective sampling method for object detection. Finally, we describe our proposed pipeline to pick up objects from unseen classes with high confidence and then explain how we can modify their objectness loss to reduce their adverse effect to improve FSOD performance.

\subsection{Finding the Potential Proposals}
\label{sec: Find the potential proposals}

In the original Faster R-CNN network~\cite{Ren15}, the unlabeled area in an image is often treated as background objects during training. As a result, we may have  incorrect objectness scores for these regions which in turn would result in poor performance on the object detection task. As a result, the potential unlabeled objects from unseen novel classes are suppressed and can be difficult to identify as true objects. However, in order to correct the objectness of these potential objects, the first step is to locate them. We have observed empirically  that the network often has clear attention to the potential objects in the RPN layer, irrespective of them being instances of seen classes.
As an example, we have used Grad-CAM visualization~\cite{selvaraju2017grad} of the feature map of the RPN layer on two representative training images, as shown in Figure.~\ref{fig:visualization}, for R-CNN. Grad-CAM is a visual interpretation technique for convolutional neural networks (CNNs), which allows users to understand the importance of different spatial regions in an image for a given prediction by highlighting the regions that contribute the most to the final decision. This technique has been widely used in computer vision applications, such as object detection and segmentation, to improve interpretability and user understanding of CNN predictions. Although novel unlabelled objects appear in the base training images, we observe upon close inspection that the RPN layer has strong attention towards them and can predict some of them as known classes. In Figure.~\ref{figure2_a} and ~\ref{figure2_b}, the feature map of $p3$ layer clearly shows the attention of the ``chair'' (base class) and ``sofa'' (novel unseen class), but the sofa is predicted to be an instance of the base class ``car''. Similarly, in Figure.~\ref{figure2_c} and ~\ref{figure2_d}, the potential novel objects from an unseen class (``bus'') can also be seen in the feature map of $p4$ layer, where the ``bus'' is predicted as an instance of the base class ``train''. Although we have provides two samples in Figure ~\ref{fig:visualization}, we can observe this situation in many other cases. This observation serves as an inspiration to identify potential proposals that contains objects from unseen classes because some novel objects have high possibilities to be predicted as known base class. Effect of these cases on few-shot learning is more severe because they can interfere more easily with a few samples. We also alter the FSOD training phase such that the model is enabled to identify and discard these instances using contrastive learning.

\begin{figure}[t]
	\centering
    \begin{subfigure}{\linewidth}
        \centering
    	\includegraphics[width=50mm]{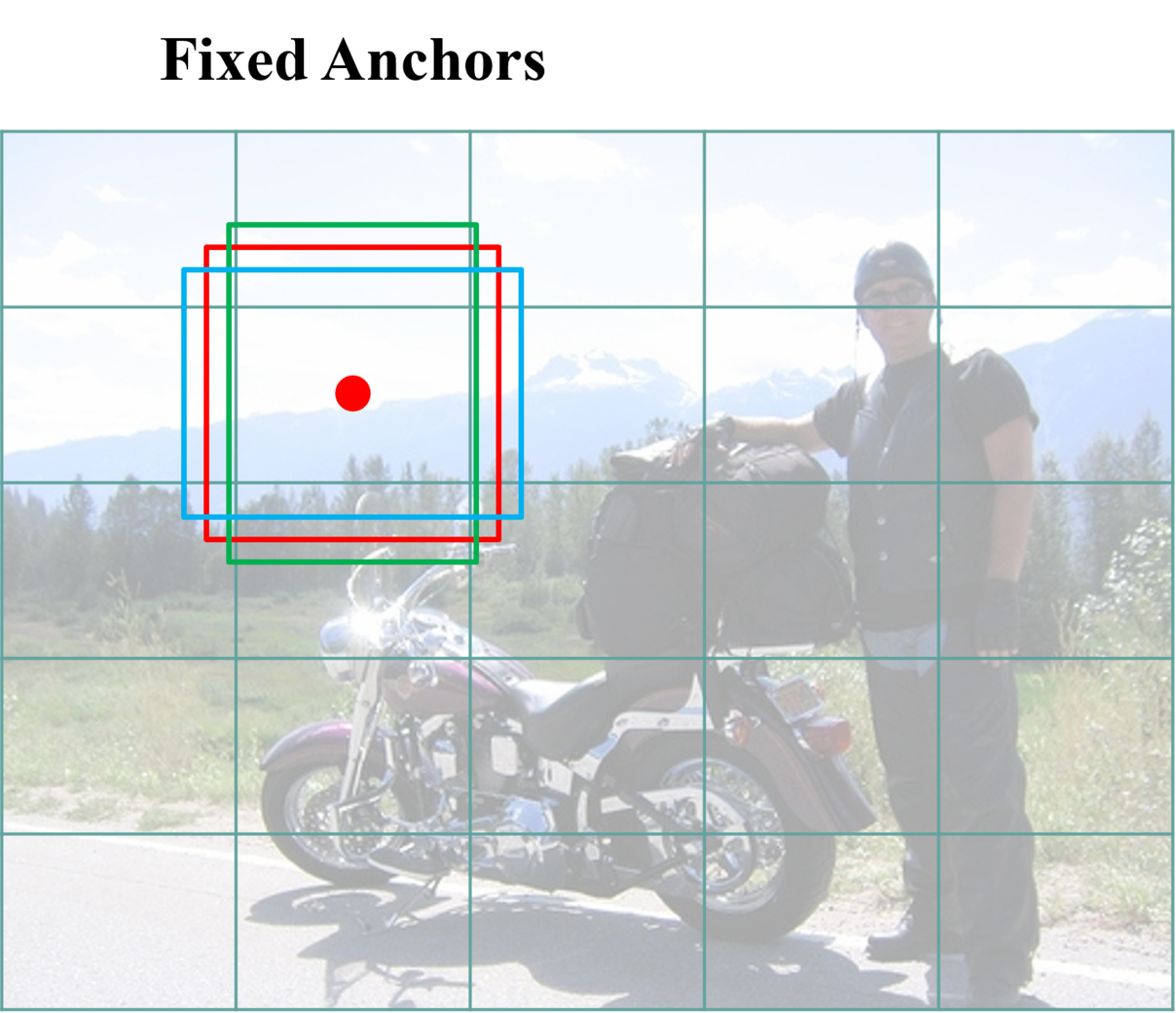}
    	\caption{}
    	\label{fig:anchors}
    \end{subfigure}
    \centering
    \begin{subfigure}{\linewidth}
        \centering
    	\includegraphics[width=125mm]{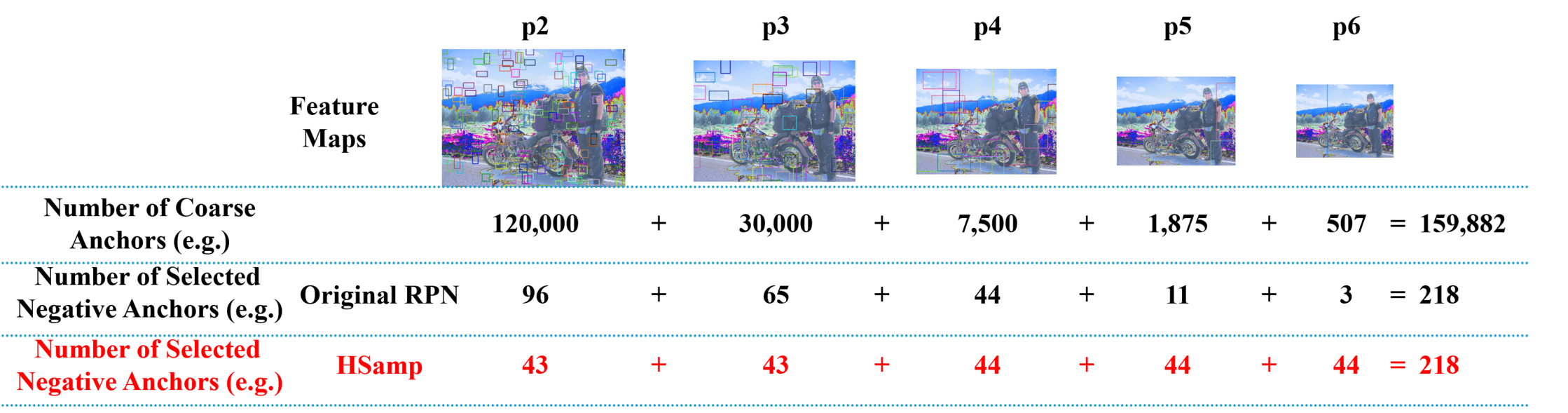}
    	\caption{}
    	\label{fig:HirecRPN}
    \end{subfigure}
	\caption{(a) A schematic diagram of how coarse anchors work: For an $n\times m$ feature map, 3 fixed anchors will be applied on each pixel of it. Therefore, each feature map would have $3m\times n$ coarse anchors. (b) The mechanism of our proposed HSamp can equally choose anchors for each layer.}
	\label{fig:HSamp}
\end{figure}

Theoretically, there is always the possibility that anchors can include potential novel objects in an image.
Unseen novel potential class objects in the base training images typically have lower $iou_{gt}^a$ ~\cite{Li21c} and therefore must be contained by negative anchors ($A_n$).  The RPN layer uses anchor boxes to determine if an area contains objects. Each pixel of the feature map is associated with three fixed anchors of different sizes and aspect ratios, as shown in Figure.~\ref{fig:anchors}. Consequently, the overall number of anchors decreases for higher dimension of feature maps. In the original RPN layer, different sizes of anchors are applied according to the size of the $p2$ to $p6$ feature maps. Larger anchors are better suited for detecting larger objects in higher feature layers because they have a larger receptive field, whereas smaller anchors are better suited for detecting smaller objects in lower feature layers. This inch-by-inch sliding window search should produce a sufficient number of candidates $A_n$, which will likely include potential unseen novel objects.
   For a better illustration, we defined all the original negative anchors as coarse negative anchors ($A_{CN}$); within $A_{CN}$, anchors contains potential novel objects are called potential novel anchors ($A_{PN}$), and others are true negative anchors ($A_{TN}$). The relationship is $A_{TN}=A_{PN}\cup A_{TN}$. Accordingly, the positive anchors are $A_P$.
To increase the training efficiency, not all anchors are employed to determine proposal boxes. For an image, only 256 $A_a$ and $A_n$ anchors among all feature maps are randomly chosen to participate in the RoI pooling. Despite its benefits, the random selection process significantly reduces the probability of obtaining desired negative anchors for large objects in higher dimension feature maps. The reason is that the anchors of larger size in $p4$ to $p6$ layers inherently have fewer possibilities.

\begin{figure*}[htbp]
	\centering
	\begin{subfigure}{0.16\linewidth}
		\centering
		\includegraphics[width=0.9\linewidth]{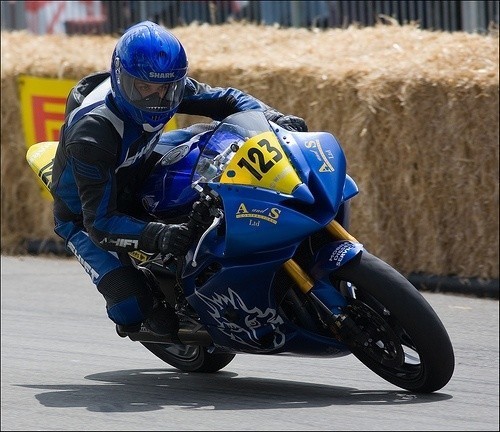}
		\caption{}
		\label{figure6_ori_a1}
	\end{subfigure}
	\centering
	\begin{subfigure}{0.16\linewidth}
		\centering
		\includegraphics[width=0.9\linewidth]{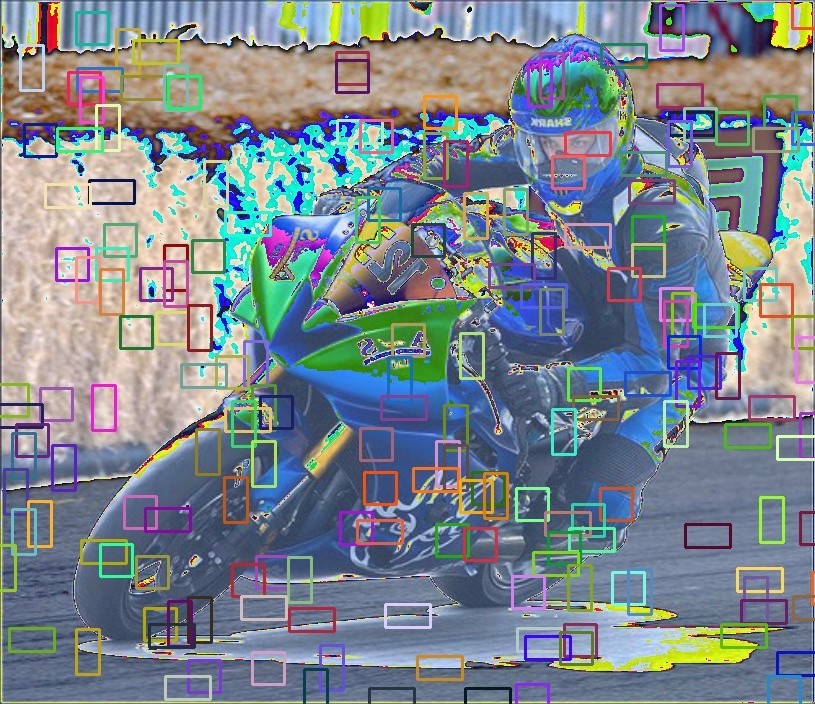}
		\caption{}
		\label{figure6_ori_a2}
	\end{subfigure}
	\centering
	\begin{subfigure}{0.16\linewidth}
		\centering
		\includegraphics[width=0.9\linewidth]{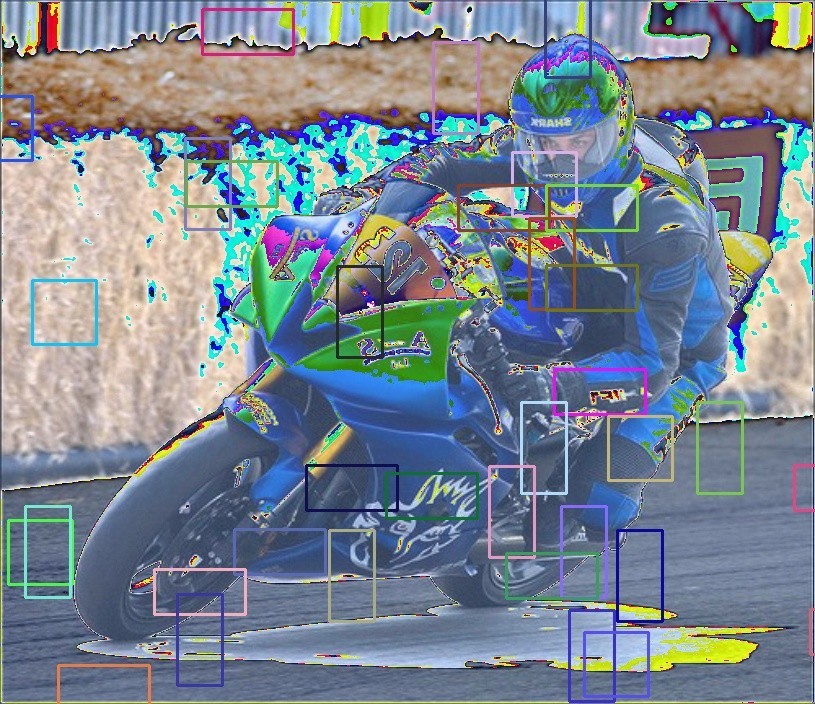}
		\caption{}
		\label{figure6_ori_a3}
	\end{subfigure}
	\centering
	\begin{subfigure}{0.16\linewidth}
		\centering
		\includegraphics[width=0.9\linewidth]{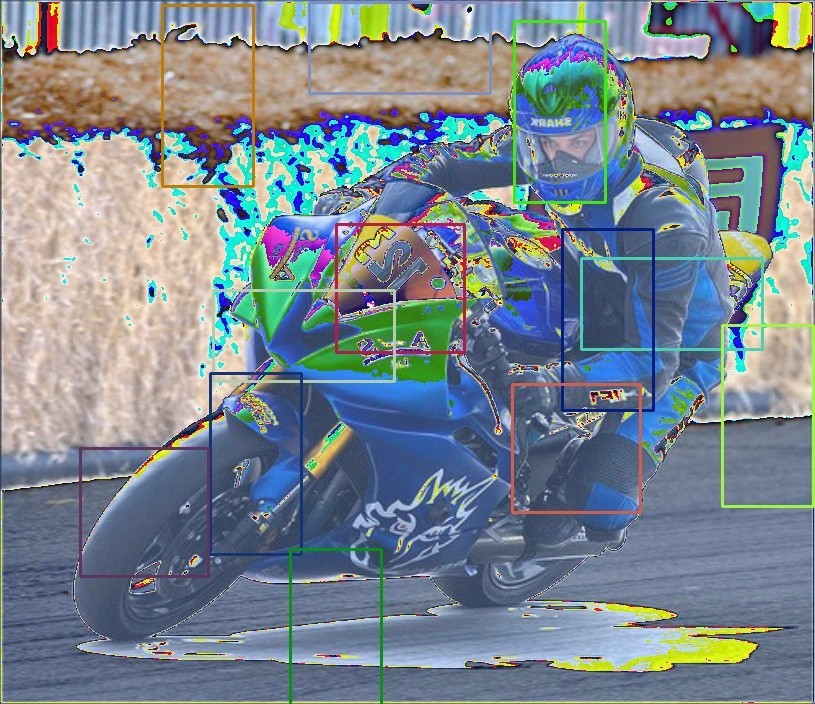}
		\caption{}
		\label{figure6_ori_a4}
	\end{subfigure}
	\centering
	\begin{subfigure}{0.16\linewidth}
		\centering
		\includegraphics[width=0.9\linewidth]{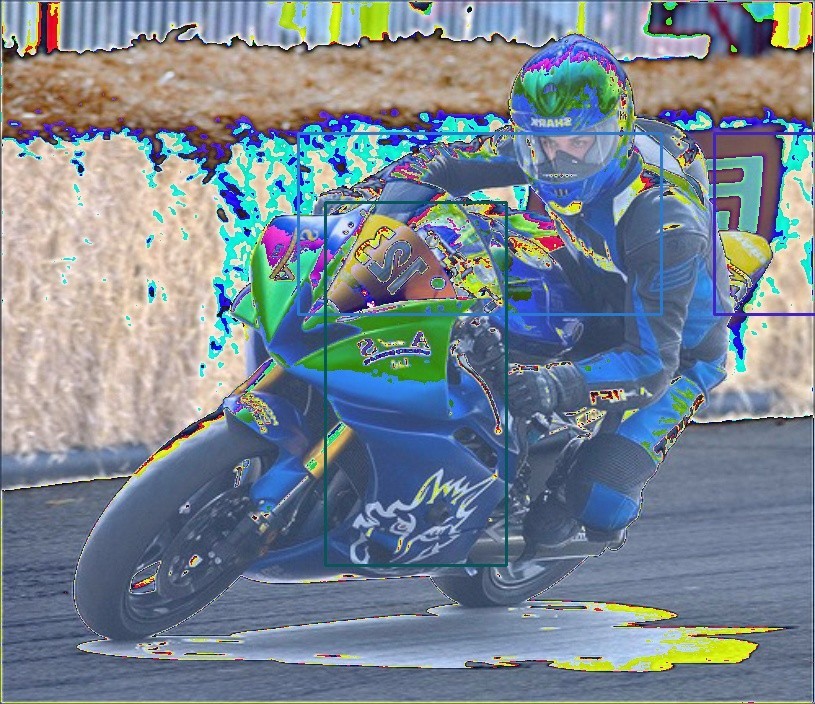}
		\caption{}
		\label{figure6_ori_a5}
	\end{subfigure}
	\centering
	\begin{subfigure}{0.16\linewidth}
		\centering
		\includegraphics[width=0.9\linewidth]{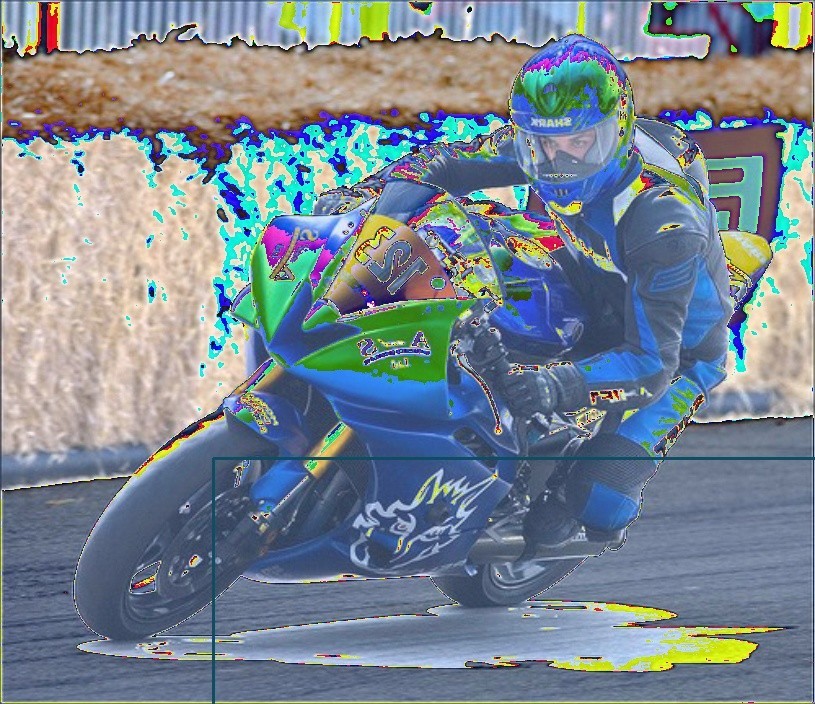}
		\caption{}
		\label{figure6_ori_a6}
	\end{subfigure}
        \centering
	\begin{subfigure}{0.16\linewidth}
		\centering
		\includegraphics[width=0.9\linewidth]{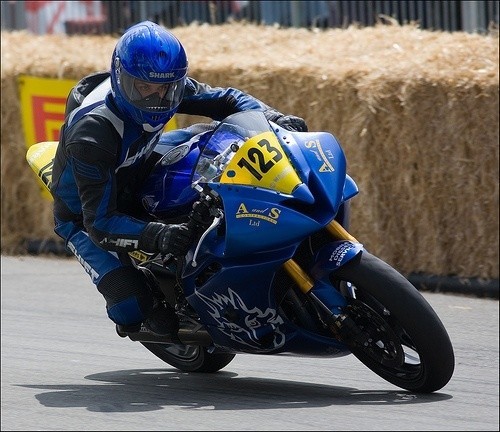}
		\caption{}
		\label{figure6_new_a1}
	\end{subfigure}
	\centering
	\begin{subfigure}{0.16\linewidth}
		\centering
		\includegraphics[width=0.9\linewidth]{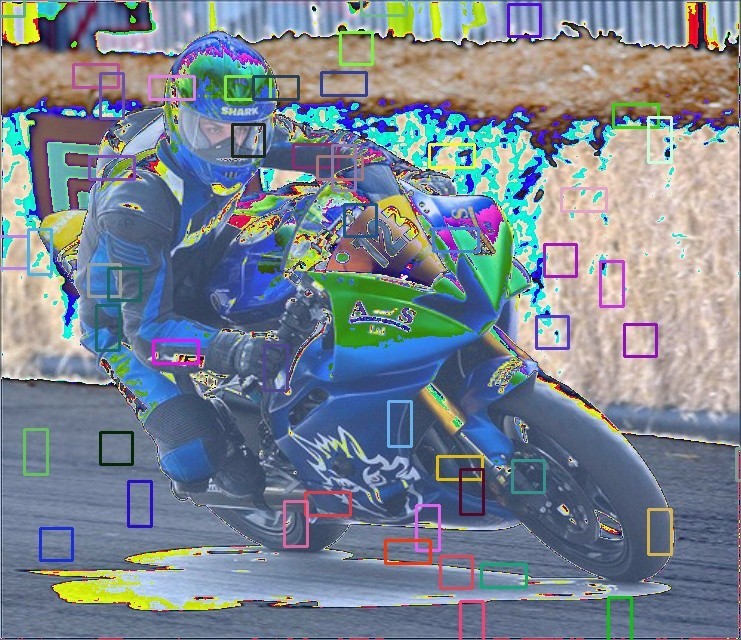}
		\caption{}
		\label{figure6_new_a2}
	\end{subfigure}
	\centering
	\begin{subfigure}{0.16\linewidth}
		\centering
		\includegraphics[width=0.9\linewidth]{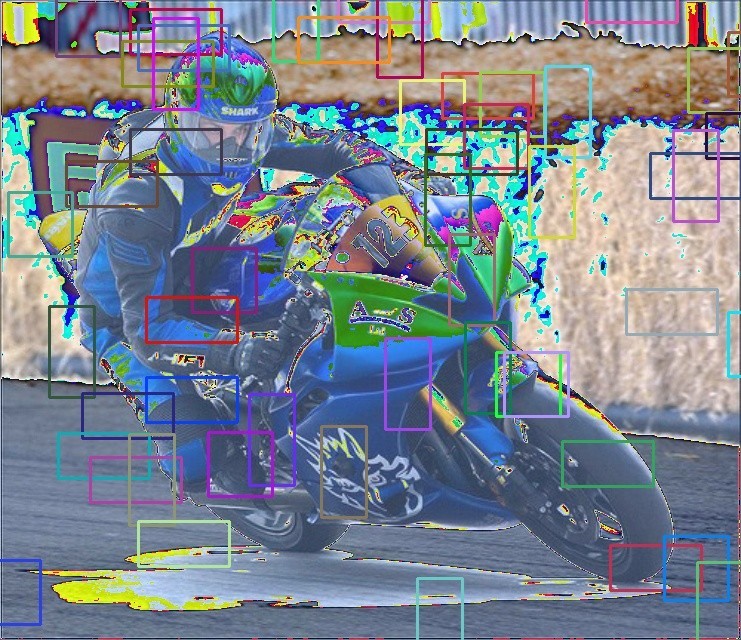}
		\caption{}
		\label{figure6_new_a3}
	\end{subfigure}
	\centering
	\begin{subfigure}{0.16\linewidth}
		\centering
		\includegraphics[width=0.9\linewidth]{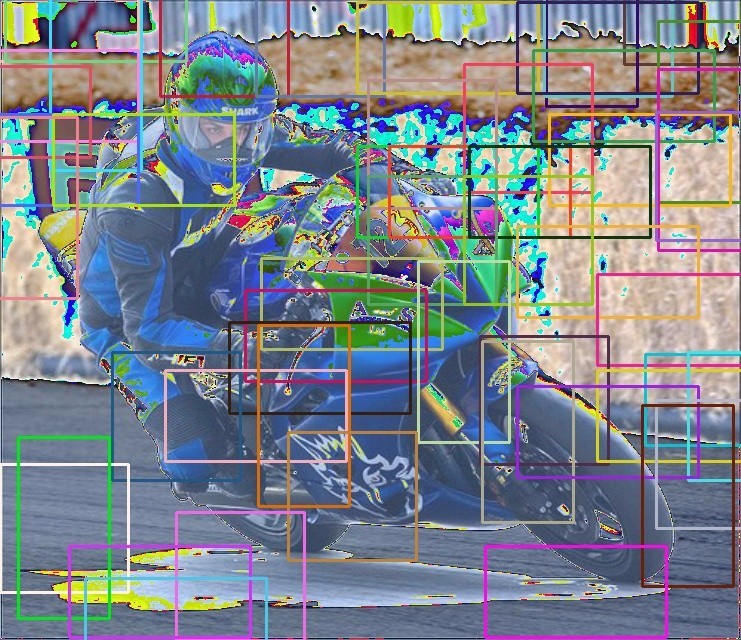}
		\caption{}
		\label{figure6_new_a4}
	\end{subfigure}
	\centering
	\begin{subfigure}{0.16\linewidth}
		\centering
		\includegraphics[width=0.9\linewidth]{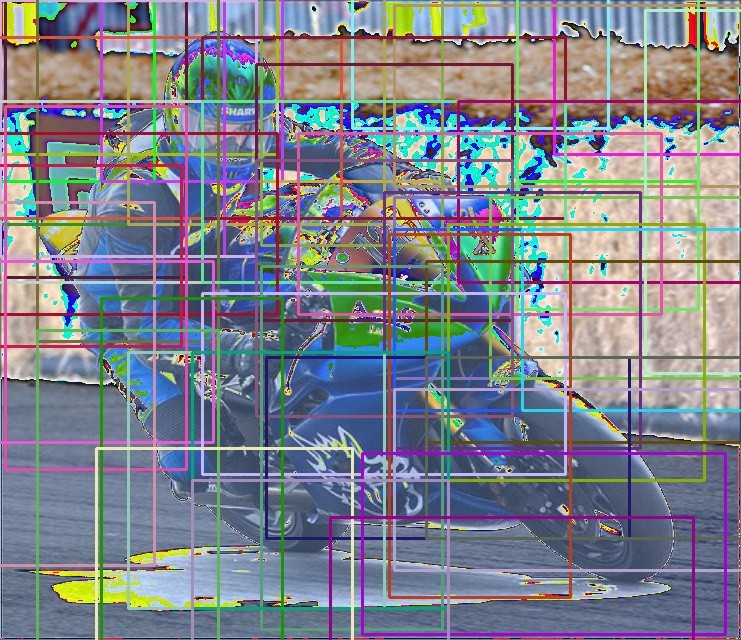}
		\caption{}
		\label{figure6_new_a5}
	\end{subfigure}
	\centering
	\begin{subfigure}{0.16\linewidth}
		\centering
		\includegraphics[width=0.9\linewidth]{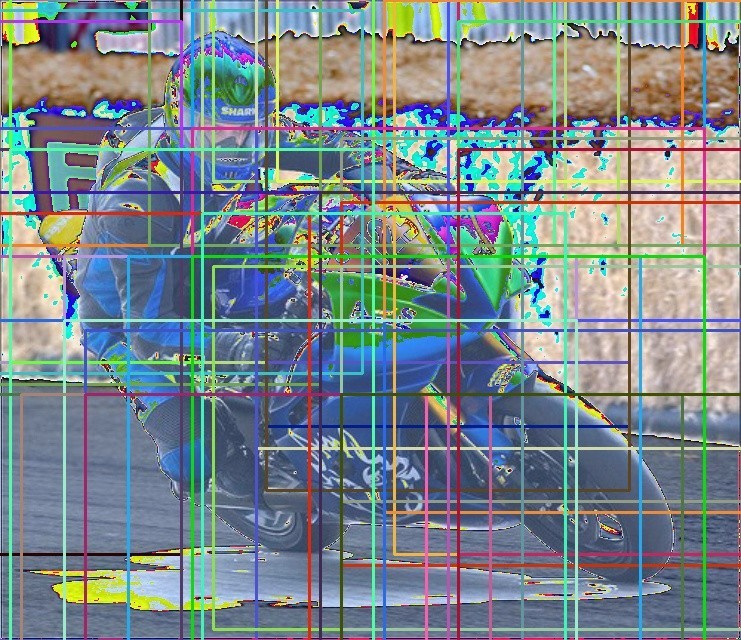}
		\caption{}
		\label{figure6_new_a6}
	\end{subfigure}
	\caption{An original image and anchors of p2/p3/p4/p5/p6 layers using the original RPN and HSamp sampling strategies: (a)-(f) in the FRCN architecture, we hardly have chance to pick up negative anchors that contain large novel unseen class objects. (g)-(l) when HSamp is used, the anchors are more balanced for each layer and the large objects are more likely to be contained and identified.}
	\label{fig:visualized HSamp}
\end{figure*}

The first novelty that we have is to adopt a new sampling strategy for anchor selection which increases the likelihood of identifying object instances from unseen classes.
To identify instances of novel unseen classes, we randomly select $A_n$s in a hierarchically balanced way, namely hierarchically sampling (HSamp). That is, if we need to pick up $m$ negative anchors ($m\textless 256$), we equally assign them to each feature layer so that each layer will have around $m/5$ anchors. By using this balanced strategy, the anchors in each feature layer would share the same possibility for being selected, as shown in Figure.~\ref{fig:HirecRPN}. Therefore, the anchors that belong to the $p4$ to $p6$ layers are safely preserved. For example, in Figure.~\ref{fig:HirecRPN}, there are [[120,000], [30,000], [7,500], [1,875], [507]] anchors for $p4$ to $p6$ layers in a training batch, and 218 negative anchors are needed. With the original RPN, these 218 $A_n$ anchors are randomly selected which means the number of $A_n$ for $p6$ feature map is only 3. In contrast, when HSamp is used, the number of $A_n$ is equal for each feature map. We have also visualized the effect of our method in Figure~\ref{fig:visualized HSamp}. There is hardly any $A_n$ that contains the motorbike (novel unseen object) when using the original RPN sampling strategy, as shown in Figure.~\ref{figure6_ori_a1} to~\ref{figure6_ori_a6}. However, when HSamp is used, the chance to have an $A_n$ that contains the motorbike is higher, as shown in Figure.~\ref{figure6_new_a1} to~\ref{figure6_new_a6}. HSamp sampling strategy is effective when using a pre-defined number of training anchors compared to the random selecting strategy. Increasing the number of training anchors is beneficial, but it also increases computational load. If more anchors are randomly selected, there may still be a \textbf{quantity bias} towards small objects over larger ones, leading to a large portion of middle to large objects being discarded. HSamp strategy addresses these shortcomings. We conclude that it is essential to employ a balanced approach in sampling negative anchors across all feature layers to identify potential novel objects that are not yet present in the seen novel classes. This method will assist us in distinguishing unseen novel class instances from those that are similar to the seen classes. We can use them to enable the model to distinguish them from instances of base classes.

\subsection{Hierarchical ternary classification region proposal network (HTRPN)}
\label{sec4.2}
As mentioned in Section~\ref{sec: Find the potential proposals}, Faster R-CNN is capable of identifying a significant number of potential novel objects that belong to unseen classes, even though they are not labeled during the training phase on the base class images. We hypothesize that this ability is due to the similarities of the feature representations between some novel unseen classes and some of the base classes. The expamples provided in Figure \ref{fig:visualization} also support this hypothesis. Consequently, the model may predict a novel unseen class object as a base class due to its resemblance to the base classes. In other words, the novel objects contained by the negative anchors may have a relatively high classification score towards a base class that is the most similar to them. We mark the negative anchors that contain potential novel unseen class objects as potential anchors ($A_p$), while others are marked as true negative anchors ($A_n^t$). Our goal is to develop a method to enable distinguishing between these two subsets. As a result, it will be less likely for the model to confuse unseen novel classes with seen classes.

\begin{figure*}[ht]
	\hfill
	\begin{subfigure}{0.9\linewidth}
		\centering
		\includegraphics[width=120mm]{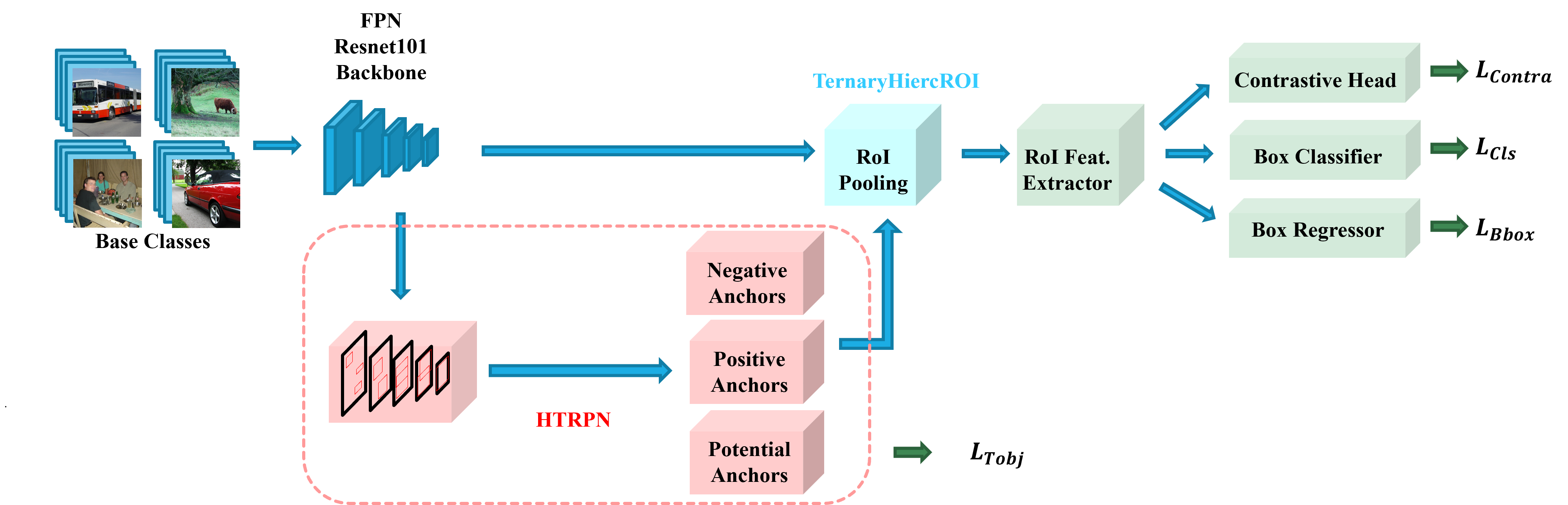}
		\caption{End-to-end architecture.}
		\label{fig:flowchart-a}
	\end{subfigure}
	\hfill
	\begin{subfigure}{0.9\linewidth}
		\centering
		\includegraphics[width=120mm]{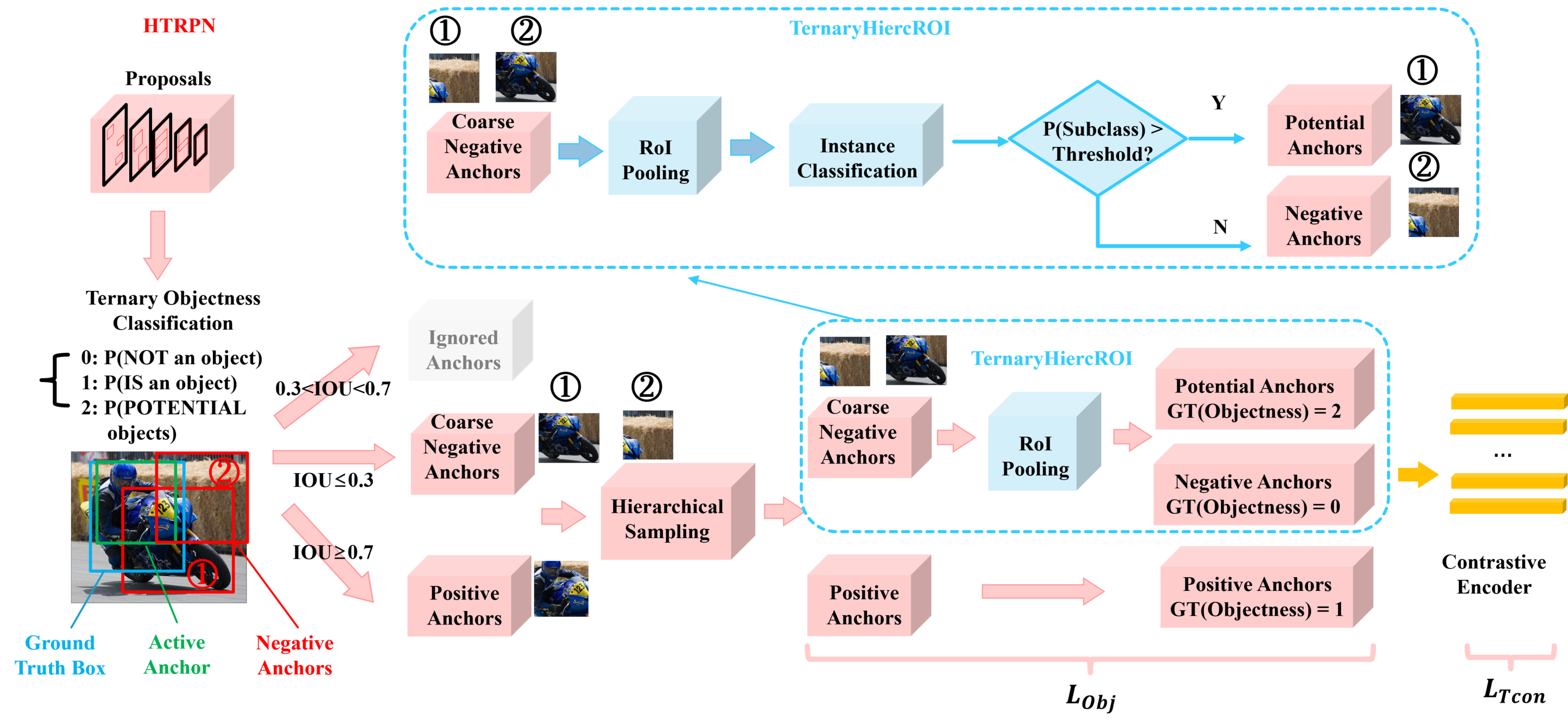}
		\caption{Detailed description of our proposed TernaryHiercRPN.}
		\label{fig:flowchart-b}
	\end{subfigure}
	\caption{The proposed TernaryHiercRPN architecture for distinguish instance of unseen classes from the seen classes: we develop a semi-supervised structure that rather solving a binary classification problem, performs ternary classification for the region proposal network that helps distinguish the potential novel objects. Note that the HSamp strategy is applied to achieve more balanced sampling for different scales of proposals.}
	\label{fig:flowchart}
\end{figure*}

Figure~\ref{fig:flowchart} visualizes our proposed architecture for improving FSOD. To better distinguish the set of $A_p$ from the set of $A_n^t$ during training, instead of performing binary classification to determine objectness in the original RPN, we propose a ternary objectness classification (i.e., $tobj\{{tobj_{pre}}^i, tobj_{gt}, iou_{gt}^a\}$, where ${tobj_{pre}}^i$ are the predicted ternary objectness scores between 0 and 1 for each class $i$; ground-truth value $tobj_{gt}=0$ indicates non-object, $tobj_{gt}=1$ represents a true object, $tobj_{gt}=2$ represents potential objects from unseen novel classes) so that potential objects that belong to unseen novel classes can be classified as a separate class, as visualized in Figure.~\ref{fig:flowchart-b}. For a training image, after hierarchically sampling the coarse negative anchors, we retain these negative anchors for each batch of anchors and perform instance-level sub-classification on them. Here, we set an instance-level classification threshold ($Thre_{cls}$). If we observe that the classification score is larger than the threshold  ($P(cls) > Thre_{cls}$) for a base class, then we set the anchor as an objectness-positive anchor, and mark its objectness loss with the ground truth of the label 2. For example the motorbike in Figure.~\ref{fig:flowchart-b}, the blue box is the ground truth box, active anchors are in green boxes, while negative anchors are in red boxes. Features of the negative anchors are sent to the RoI pooling layer to see if they could be predicted as a visually similar seen category (e.g. the negative anchor \ding{172} is predicted as base class ``bicycle'', then it is assigned with $tobj_{gt}=2$; but for anchor \ding{173} is kept as $tobj_{gt}=0$ since it does not pass the $Thre_{cls}$. We argue that our novel architecture will have a higher FSOD performance.

\begin{figure}[t]
    \centering
    \includegraphics[width=100mm]{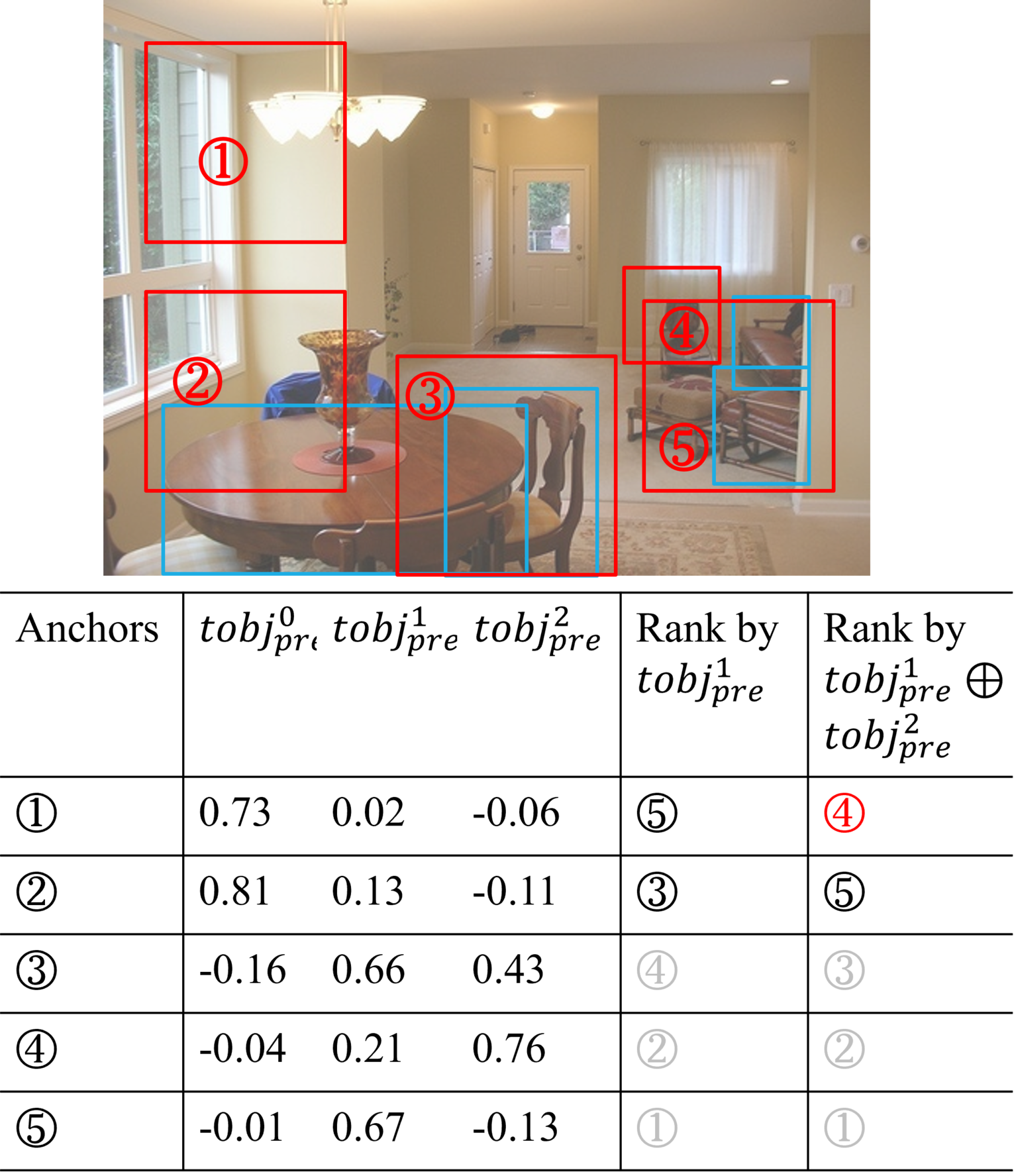}
    \caption{ \small The 4 blue boxes are the ground truth boxes and the 5 red boxes are the proposals. The proposal \ding{175} contains an unlabeled novel object chair. The top 2 proposals that are ranked by ${tobj_{pre}}^1 \oplus {tobj_{pre}}^2$ could successfully include proposal \ding{175}.}
    \label{fig:combine}
\end{figure}

Furthermore, when implementing our proposed HTRPN, we also need tailored solutions for pre-training and fine-tuning. Due to computational resource constraints, only the top 1000 proposals are utilized for RoI pooling in conventional RPNs. During the pre-training phase, proposals are ranked based on their objectness scores  ${tobj_{pre}}^1$ since the model is solely trained to recognize the base classes at this stage. However, in the fine-tuning stage, the ${tobj_{pre}}^1$ and ${tobj_{pre}}^2$ are both considered for ranking the proposals, because the objectness of some labeled objects might be predicted as ${tobj_{pre}}^2$ due to knowledge transfer from the pre-training stage. This step is crucial to realize the objectness consistency because the combination of ${tobj_{pre}}^1$ and ${tobj_{pre}}^2$ could represent the highly confident proposal and especially improve the possibility of determining positive anchors while inferencing. As shown in Figure~\ref{fig:combine}, if the top two proposals out of the five proposals are ranked only using ${tobj_{pre}}^1$, then the proposal \ding{175} would be ignored. However, when the top two proposals are ranked by ${tobj_{pre}}^1 \oplus {tobj_{pre}}^2$, the proposal \ding{175} could be correctly included. This scheme significantly enhances the likelihood of discovering true objects, allowing for a more accurate distinction between potential novel unseen class objects and coarse negative anchors. Consequently, the ternary RPN enables the model to maintain its ability to identify novel objects from previously unseen classes. In practice, not all potential objects that exist in training datasets are necessarily identified during training and only a subset of them may be detected. Nonetheless, the novel unseen class objects that have been successfully isolated by this process can still help to reduce confusion in few-shot learning due to their dissimilarity to the known classes.
We use these identified objects along with contrastive learning to equip the model with the capability of distinguishing them.

\subsection{Contrastive Learning on Objectness}

We noted that sample imbalance can be particularly significant in the RPN layer. To investigate this possibility, we calculated the number of foreground and background proposals generated by the RPN of a standard Faster R-CNN network, and the result indicates that the ratio of foreground over background proposals is around 0.17. This number indicates that  the background proposals dominate the objectness loss. Consequently, proposals are more likely to be classified as background. Therefore, we suppose that a contrastive learning loss is necessary to address the challenge of objectness confusion. To further increase the inter-class distances between $A_a$, $A_n^t$, and $A_p$ subsets in HTRPN, we   include an objectness contrastive learning head in our FSOD method~\cite{liu2021learning,gao2021contrastive,ouali2021spatial}. Inspired by the existing literature \cite{Sun21,Khosla20}, the cropped features of proposals are sent into an encoder with their ground truth objectness logits to perform contrastive learning. The features of proposals are encoded as a 128-dimensional default feature vector, and then cosine similarity scores are calculated between each pair of proposals to compute the contrastive learning loss. Those sample pairs with higher cosine similarity that do not belong to the same category will need to have a higher contrastive loss value, and vice versa, to enforce making them more distant in the feature space. In this manner, the HTRPN would assign higher objectness scores to the proposals.

\subsection{Training Loss}

Our global training loss is composed of classic FSOD loss terms and new loss terms to implement our ideas.
It is composed of the classification loss ($\mathcal L_{Cls}$) to guide object detection, the bounding box regression loss ($\mathcal L_{Bbox}$) to guide localization, our ternary objectness loss ($\mathcal L_{Tobj}$) to enable identifying instances of unseen classes, and the RoI feature contrastive loss $\mathcal L_{Contra}$, as follows:
\begin{equation}
\mathcal L = \mathcal L_{Cls} + \mathcal L_{Bbox} + \mathcal L_{Tobj} + \alpha \mathcal L_{Contra}.
\label{eq:loss 1}
\end{equation}
The first two terms are standard terms in FSOD literature and the loss $\mathcal L_{Contra}$   is computed using the contrastive head as described in FSCE ~\cite{Sun21}. We set $\alpha = 0.5$ to be the fixed weight for balancing the contrastive learning loss based on a cross validation study.

Our proposed ternary objectness loss $\mathcal L_{Tobj}$ in Equation.~\ref{eq:loss 2} is a sum of the cross entropy objectness loss ($\mathcal L_{Obj}$) and ternary RPN feature contrastive learning loss ($\mathcal L_{Tcon}$) terms. Similar to $\alpha$ in Equation.~\ref{eq:loss 1}, $\lambda$ is a balancing factor that is set to be equal 0.5 in our experiments.  

\begin{equation}
\mathcal L_{Tobj} = \mathcal L_{Obj} + \lambda \mathcal L_{Tcon}
\label{eq:loss 2}
\end{equation}

The ternary RPN contrastive learning loss $\mathcal L_{Tcon}$, is defined as an arithmetic mean of the weighted supervised contrastive learning loss $\mathcal L_{z_i}$ as the following:
\begin{equation}
\mathcal L_{Tcon}=\frac{1}{N_{Prop}}\sum_{i=1}^{N_{Prop}}w(iou_{gt}^p)\cdot\! \mathcal L_{z_i},
\label{eq:loss 3}
\end{equation}
 where $N_{Prop}$ represents the number of RPN proposals. Weights $w(iou_{gt}^p)$ are also assigned by the function $g(\cdot)$ as:
\begin{equation}
w(iou_{gt}^p) = \mathcal I\{iou_{gt}^p\geq \phi\}\cdot g(iou_{gt}^p),
\label{eq:loss 4}
\end{equation}
where $g(\cdot)$ is a good hard-clip function ~\cite{Sun21} and $\mathcal I\{\cdot\}$ is a cut-off thresholding function that is   1 when $iou_{gt}^p\geq \phi\ $, otherwise is 0.

Finally, $\mathcal L_{z_i}$ in the RPN proposal contrastive learning loss is given as:
{\small
\begin{equation}
\mathcal L_{z_i}=\!\frac{-1}{N_{obj_{gt}^i}\!-\!1}\!\sum_{j=1,j\neq i}^{N_{Prop}}\!\mathcal I\{obj_{gt}^i= obj_{gt}^j\}\cdot\log\!\frac{e^{\tilde{z_i} \cdot\!\tilde{z_j}\!/\!\tau}}{\sum\limits_{k=1}\limits^{N_{Prop}}\!\mathcal I_{k\neq i}\cdot e^{\tilde{z_i}\cdot\tilde{z_k}/\!\tau}},
\label{eq:loss 5}
\end{equation}
}
where $z_i$ denotes the contrastive feature,  $obj_{gt}^i$ denotes the ground truth ternary objectness label for the $i$-th proposal,   $\tilde{z_i}$ denotes normalized features while measuring the cosine distances, and $N_{obj_{gt}^i}$ denotes the number of proposals with the same objectness label as $obj_{gt}^i$.

\section{Experimental Results}

We demonstrate the effectiveness of our proposed architecture and training procedure in enhancing the performance of FSOD. Our implementation is publicly available to ensure reproducibility: ~\url{https://github.com/zshanggu/HTRPN}. As a result, all details are  self-contained in the codebase to facilitate future research and comparison.

\subsection{Experimental Setup}

\textbf{Object detection model and the evaluation protocol:} As our object detection model, we employ Faster R-CNN, which is supported by a ResNet-101 backbone and a feature pyramid network (FPN)~\cite{lin2017feature}. The evaluation scheme strictly follows the same paradigm as described in TFA~\cite{Wang20} to enable comparison with results reported in other works on FSOD. To ensure a fair comparison, when training the base class, we adhere to the TFA~\cite{Wang20} and FSCE~\cite{Sun21} guidelines and utilize the official ImageNet pre-trained model.  The contrastive learning head in the fine-tuning stage is computed similarly to FSCE~\cite{Sun21}.  In the evaluation protocol, a two-fold approach is employed to assess the performance of our model. Specifically, we calculate the mean average precision (mAP50) for each category in the dataset, using both the base classes (bAP50) and the novel seen class (nAP50). The AP50 value indicates the average precision (AP) at an IoU (Intersection over Union) threshold of 0.5, which means that the model detects objects with an IoU of 0.5 or higher from the ground truth. We repeat the experiments five times and report the average values.

\textbf{Datasets:} the results are reported on the PASCAL VOC~\cite{everingham2010pascal,everingham2015pascal} and COCO~\cite{lin2014microsoft} datasets. These datasets have been extensively used for evaluating the performance of few-shot object detection models and have become a standard benchmark for researchers in this area~\cite{Wang20}. 

\textbf{Optimization  setup:} The optimizer is fixed as SGD and the weight decay is 1e-4 with momentum as 0.9. We set our batch size equal to 16 for all experiments. The $Thre_{cls}$ is fixed as 0.75. These hyperparameters are not fine-tuned.  We also provide an empirical  study about tuning the hyper-parameters for optimal performance.

In the pre-training stage, the top 1000 proposals, used for RoI pooling, are ranked by the second objectness logit (is an object). While in the fine-tuning stage, the top 1000 proposals are ranked by the maximum of the second and the third objectness logits (potential object). 

\textbf{Baselines for comparison:} There are many existing FSOD methods. We compare our performance against a subset of recently developed SOTA FSOD methods to demonstrate that our method os competitive against recent developments in the feild. These methods include: LSTD  ~\cite{Chen18}, YOLOv2-ft ~\cite{Wang19}, RepMet  ~\cite{Karlinsky19}, FRCN  ~\cite{Wang19},  TFA   ~\cite{Wang20}, MPSR~\cite{Wu20}, Retentive R-CNN ~\cite{Fan21}, FSCE  ~\cite{Sun21}, TIP~\cite{Li21a}, DC-Net~\cite{Hu21a}, FSOD-UP  ~\cite{Wu21b}, CME~\cite{Li21b}, KFSOD   ~\cite{Zhang22}, 
SRR-FSD ~\cite{Zhu21}, SVD ~\cite{Wu21a}, FORD+BL ~\cite{Nguyen22}, and N-PME ~\cite{Liu22}. The performance of  each method is reported if the original paper provides experiments on the dataset we used.

\textbf{Implemented details:}
We use 4 Nvidia T4 Tensor Core GPUs for both training and evaluation of all datasets. The general hyperparameters we used are listed in Table.~\ref{tab:hyperparameters}. Due to the differences in feature spaces between pre-training and fine-tuning samples, the weights of the contrastive learning modules used for RPN and RoI pooling cannot be transferred from the pre-trained model to the fine-tuning tasks. Therefore, we remove these weights from the pre-trained model and unfreeze the weights of the ResNet backbone and RoI pooling during fine-tuning training.

\begin{table}[t]
	\centering
	\resizebox{0.75\textwidth}{!}{
	\begin{tabular}{l|l|l}
		\toprule
		  Hyperparameters & PASCAL VOC & COCO \\
		\midrule
		Learning rate & 0.01 & 0.001 \\
		Weight decay & 1e-4 & 1e-4 \\
		Optimizer & SGD & SGD \\
		Temperature & 0.2 & 0.2 \\
		Contrastie loss weight & 0.5 & 0.5 \\
		$Thre_{cls}$ & 0.75 & 0.75 \\
		\bottomrule
	\end{tabular}
    }
	\caption{Hyper parameters that we use for experiments.}
	\label{tab:hyperparameters}
\end{table}

\subsection{Comparison Results}

\subsubsection{Results on PASCAL VOC}

The PASCAL VOC   dataset is   commonly used in computer vision   for  object detection. 
It contains images from a variety of real-world scenes and encompasses a diverse set of object classes, making it suitable for evaluating the performance of algorithms designed to detect and classify objects within images. Each image in the dataset is annotated with bounding boxes that indicate the location of different objects, along with labels that specify the object's class.
For FSOD using the PASCAL VOC 2007 and 2012 datasets, a set of 15 categories has been designated as the base classes for pre-training purposes, while the remaining 5 categories are designated as the  novel  seen classes. The division of categories aligns with the three distinct category splits introduced in the TFA  \cite{Wang20}. To ensure a fair comparison, Wang et al.\cite{Wang20} have defined three distinct combinations of the base and the novel seen classes, referred to as split1, split2, and split3.
Across each of these category splits, we conduct evaluations by measuring the average precision for novel classes (referred to as nAP) at different few-shot learning levels, namely 1, 2, 3, 5, and 10 shots. The training iterations comprises 8000 iterations per training epoch, with an initial learning rate set at 0.02. 

The outcomes of our experiments are tabulated in Table~\ref{tab:voc}.  We observe that no FSOD methods has the best performance under all situations and a FSOD method should be deemed a good method if it leads to a competitive performance across all cases.
Upon   scrutiny of the results, we   observe that the results for split3 surpass those of split1 and split2. Upon careful examination of the images, we think that this observation can be attributed   to the relative simplicity of the background in split3. The backgrounds in split3 exhibit reduced complexity, rendering them more amenable to our analysis compared to the intricate backgrounds encountered in split1 and split2 that makes FSOD less vulnerable with respect to instances of unseen novel classes.
Furthermore, we undertake a thorough comparative study between our obtained results and an in-house implementation of the FSCE. Our empirical investigation reveals that our proposed method has outperformed this method, implying that our approach is not only more efficacious but also more resource-efficient compared to the original FSCE method.
In conclusion, we find that our proposed approach consistently enhances performance across a variety of scenarios. Notably, our method demonstrates pronounced efficacy when dealing with scenarios involving a smaller number of shots ($n$ shots).

\begin{table*}[ht]
	\centering
	\resizebox{\textwidth}{!}{
		\begin{tabular}{ll|l|lllll|lllll|lllll}
			\toprule
			\multicolumn{2}{l|}{\multirow{2}{*}{\diagbox[height=25pt,innerrightsep=37pt]{Method}{Shot}}} & \multirow{2}{*}{Backbone} & \multicolumn{5}{c|}{Split1} & \multicolumn{5}{c|}{Split2} & \multicolumn{5}{c}{Split3} \\
			& & & 1 & 2 & 3 & 5 & 10 & 1 & 2 & 3 & 5 & 10 & 1 & 2 & 3 & 5 & 10\\
			\midrule
			LSTD & AAAI 18 ~\cite{Chen18} & \multirow{2}{*}{VGG-16} & 8.2 & 1.0 & 12.4 & 29.1 & 38.5 & 11.4 & 3.8 & 5.0 & 15.7 & 31.0 & 12.6 & 8.5 & 15.0 & 27.3 & 36.3 \\
			YOLOv2-ft & ICCV19 ~\cite{Wang19} & & 6.6 & 10.7 & 12.5 & 24.8 & 38.6 & 12.5 & 4.2 & 11.6 & 16.1 & 33.9 & 13.0 & 15.9 & 15.0 & 32.2 & 38.4 \\
			\midrule
			RepMet & CVPR 19 ~\cite{Karlinsky19} & InceptionV3 & 26.1 & 32.9 & 34.4 & 38.6 & 41.3 & 17.2 & 22.1 & 23.4 & 28.3 & 35.8 & 27.5 & 31.1 & 31.5 & 34.4 & 37.2 \\
			\midrule
			FRCN-ft & ICCV19 ~\cite{Wang19} & \multirow{12}{*}{FRCN-R101} & 13.8 & 19.6 & 32.8 & 41.5 & 45.6 & 7.9 & 15.3 & 26.2 & 31.6 & 39.1 & 9.8 & 11.3 & 19.1 & 35.0 & 45.1 \\
			FRCN+FPN-ft & ICML 20 ~\cite{Wang20} & & 8.2 & 20.3 & 29.0 & 40.1 & 45.5 & 13.4 & 20.6 & 28.6 & 32.4 & 38.8 & 19.6 & 20.8 & 28.7 & 42.2 & 42.1 \\
			TFA w/ fc & ICML 20 ~\cite{Wang20} & & 36.8 & 29.1 & 43.6 & 55.7 & 57.0 & 18.2 & 29.0 & 33.4 & 35.5 & 39.0 & 27.7 & 33.6 & 42.5 & 48.7 & 50.2 \\
			TFA w/ cos & ICML 20 ~\cite{Wang20} & & 39.8 & 36.1 & 44.7 & 55.7 & 56.0 & 23.5 & 26.9 & 34.1 & 35.1 & 39.1 & 30.8 & 34.8 & 42.8 & 49.5 & 49.8 \\
			MPSR & ECCV 20 ~\cite{Wu20} & & 41.7 & - & 51.4 & 55.2 & 61.8 & 24.4 & - & 39.2 & 39.9 & 47.8 & 35.6 & - & 42.3 & 48.0 & 49.7 \\
			Retentive R-CNN & CVPR 21 ~\cite{Fan21} & & 42.4 & 45.8 & 45.9 & 53.7 & 56.1 & 21.7 & 27.8 & 35.2 & 37.0 & 40.3 & 30.2 & 37.6 & 43.0 & 49.7 & 50.1 \\
			FSCE & CVPR 21 ~\cite{Sun21} & & 44.2 & 43.8 & 51.4 & \textcolor{blue}{61.9} & 63.4 & 27.3 & 29.5 & \textcolor{blue}{43.5} & 44.2 & 50.2 & \textcolor{blue}{37.2} & \textcolor{blue}{41.9} & \textcolor{blue}{47.5} & \textcolor{blue}{54.6} & \textcolor{blue}{58.5} \\
			TIP & CVPR 21 ~\cite{Li21a} & & 27.7 & 36.5 & 43.3 & 50.2 & 59.6 & 22.7 & 30.1 & 33.8 & 40.9 & 46.9 & 21.7 & 30.6 & 38.1 & 44.5 & 50.9 \\
			DC-Net & CVPR 21 ~\cite{Hu21a} & & 33.9 & 37.4 & 43.7 & 51.1 & 59.6 & 23.2 & 24.8 & 30.6 & 36.7 & 46.6 & 32.3 & 34.9 & 39.7 & 42.6 & 50.7 \\
			FSOD-UP & ICCV 21 ~\cite{Wu21b} & & 43.8 & \textcolor{red}{47.8} & 50.3 & 55.4 & 61.7 & \textcolor{blue}{31.2} & \textcolor{blue}{30.5} & 41.2 & 42.2 & 48.3 & 35.5 & 39.7 & 43.9 & 50.6 & 53.5 \\
			CME & CVPR 21 ~\cite{Li21b} & & 41.5 & \textcolor{blue}{47.5} & 50.4 & 58.2 & 60.9 & 27.2 & 30.2 & 41.4 & 42.5 & 46.8 & 34.3 & 39.6 & 45.1 & 48.3 & 51.5 \\
			KFSOD & CVPR 22 ~\cite{Zhang22} & & \textcolor{blue}{44.6} & - & \textcolor{red}{54.4} & 60.9 & \textcolor{red}{65.8} & \textcolor{red}{37.8} & - & 43.1 & \textcolor{red}{48.1} & \textcolor{blue}{50.4} & 34.8 & - & 44.1 & 52.7 & 53.9 \\
		\midrule
         FSCE (we implemented) & & \multirow{2}{*}{FRCN-R101} & 46.1 & 42.1 & 51.2 & 61.1 & 63.2 & 28.3 & 31.3 & 44.9 & 45.2 & 51.8 & 36.8 & 45.7 & 48.1 & 55.7 & 57.9 \\
         Ours & & & \textcolor{red}{47.0} & 44.8 & \textcolor{blue}{53.4} & \textcolor{red}{62.9} & \textcolor{blue}{65.2} & 29.8 & \textcolor{red}{32.6} & \textcolor{red}{46.3} & \textcolor{blue}{47.7} & \textcolor{red}{53.0} & \textcolor{red}{40.1} & \textcolor{red}{45.9} & \textcolor{red}{49.6} & \textcolor{red}{57.0} & \textcolor{red}{59.7} \\
			\bottomrule
		\end{tabular}
    }
	\caption{Performance results on the novel seen classes of nAP50fpr for the PASCAL VOC dataset: the methods are evaluated on three different category splits with 1 to 10-shot scenarios.  The \textcolor{red}{highest score} of each few-shot setting is in red color, and the \textcolor{blue}{second highest score} is in blue color.}
	\label{tab:voc}
\end{table*}

\subsubsection{Results on the COCO Dataset}

The Common Objects in Context (COCO) dataset initially   provides a diverse   collection of images that contain objects in complex scenes and various contexts. For object detection, it contains bounding box annotations around objects in the images, along with the class labels of the objects. For FSOD, 60 categories are selected as base classes, and the remaining 20 categories are served as novel classes. The training iterations are set to 20000 during the training stage with an initial learning rate of 0.01. AP for novel classes is evaluated upon $n=10$ and $n=30$ shots separately  using 5000 images from the 2014 COCO validation split. Our experiment results for COCO are shown in Table.~\ref{tab:coco}. We again observe that our method exceeds the performance of previous works in all cases, and in some instances, the margin of improvement is significant. These experiments provide substantive evidence for the effectiveness of our proposed approach.

For the COCO dataset, a total of 60   categories are   chosen to serve as the   base classes for our pre-training stage. The remaining subset of 20 categories is designated as the novel seen classes  for few-shot learning. Throughout the training phase, we employ   20,000 training iterations, with the initial learning rate set at 0.01 to guide the optimization process. We assess the Average Precision (AP)  under two   scenarios of  10-shot and 30-shot learning. The performance results d are   documented in Table~\ref{tab:coco}.
Once again, we observe that our proposed method consistently outperforms the  prior methods. Notably, in the case of 30-shot learning the extent of improvement achieved by our method is notably substantial which shows that our method is able to separate instances of novel unseen classes better. We conclude that our approach is helpful to improve FSOD performance.

\begin{table}[t]
	\centering
	\resizebox{0.75\textwidth}{!}{
		\begin{tabular}{ll|ll|ll}
			\toprule
			\multicolumn{2}{l|}{\multirow{2}{*}{\diagbox[innerrightsep=30pt]{Method}{Shot}}}&\multicolumn{2}{c|}{Novel AP}&\multicolumn{2}{c}{Novel AP75}\\
			&&10&30&10&30\\
			\midrule
			TFA w/ cos&ICML20~\cite{Wang20}&10.0&13.7&9.3&13.4\\
			FSCE&CVPR21~\cite{Sun21}&11.9&\textcolor{blue}{16.4}&\textcolor{blue}{10.5}&\textcolor{blue}{16.2}\\
			SRR-FSD&CVPR21~\cite{Zhu21} &11.3&14.7&9.8&13.5\\
			SVD&NeurIPS21~\cite{Wu21a}&\textcolor{blue}{12.0}&16.0&10.4&15.3\\
			FORD+BL&IMAVIS22~\cite{Nguyen22}&11.2&14.8&10.2&13.9\\
			N-PME&ICASSP22~\cite{Liu22}&10.6&14.1&9.4&13.6\\
			\midrule
			Our& &\textcolor{red}{12.1}&\textcolor{red}{17.2}&\textcolor{red}{11.2}&\textcolor{red}{17.1}\\
			\bottomrule
		\end{tabular}
	}
	\caption{Evaluation on COCO dataset for novel classes for AP and AP75 settings. The \textcolor{red}{highest score} of each few-shot setting is in red, and the \textcolor{blue}{second highest score} is in blue.}
	\label{tab:coco}
\end{table}

\subsection{Ablative and Analytic Experiments}
We provide additional  experiments to offer a better understanding about the proposed method and provide intuition behind its improved performance.

\subsubsection{Ablation Study}

We evaluate the individual contributions of our proposed modules of the proposed approach on the downstream performance, including, the hierarchically sampling mechanism of the ternary RPN, the objectness classification, and the contrastive head of the objectness. To this end, we conducted an ablation study experiment on the PASCAL VOC dataset using the 5-shot setting. This study helps analyzing the impact of each component and understand their overall contribution to the model's performance. Each proposed module is sequentially added to the base network in a cumulative manner, allowing us to analyze their individual and combined effects on the overall performance of the network. The results are presented in Table.~\ref{tab:ablation1}. We observe that all our proposed modules are necessary for optimal performance. By adding the HSamp, we can see that a balanced sampling in RPN is necessary, as it provides comprehensive improvement of both $bAP$ and $nAP$ during the pre-training and the fine-tuning stages. We can also observe the results of adding the ternary objectness module indicate that our method will further improve the $nAP$ and do no significant harm to the $bAP$. This observation is expected since we learn more class-agnostic information beyond the base classes when this module is added because the ability of recognizing potential novel objects brings extra positive anchors which interfere with detecting the base classes but as demonstrated is our results, the commutative effect is improvement of FSOD performance. While the contrastive objectness part demonstrated that it is a simple yet effective way to help build a stronger RPN that could further improve the $bAP$ and $nAP$.

\begin{table}[t]
	\centering
	\resizebox{0.75\textwidth}{!}{
	\begin{tabular}{l|l|l|l}
		\toprule
		  Modules & \makecell[l]{bAP\\(pre-trained)} & \makecell[l]{bAP\\(fine-tuned)} & \makecell[l]{nAP\\(fine-tuned)} \\
		\midrule
		FSCE Baseline* & 80.5 & \textbf{68.9} & 57.2 \\
		+ HSamp & \textbf{80.7} & \textbf{68.9} & 57.6 \\
		+ Ternary Objectness & 78.5 & 67.8 & 61.9 \\
		+ Contrastive Objectness & 78.9 & 68.6 & \textbf{62.9} \\
		\bottomrule
	\end{tabular}
    }
	\caption{Ablation studies on different modules of the proposed method. The effect of incrementally adding each module to the Baseline network is demonstrated respectively. Sign * represents our reproductive results. We listed the base class mAP50 (bAP) during the pre-training and fine-tuning stage, as well as the novel class mAP50 (nAP) during  fine-tuning.}
	\label{tab:ablation1}
\end{table}

Additionally, we study the influence of different hyperparameter $Thre_{cls}$ settings on the downstream performance. We use five $Thre_{cls}$ values from 0.05 to 0.95 for training and record the $nAP$ accordingly, as shown in Table.~\ref{tab:ablation2}. We observe that for lower $Thre_{cls}$, more  potential novel proposal candidates can be distinguished. However, we have lower confidence and consequently lower quality because many proposal canidates are irrelevant. In contrast, when a higher threshold $Thre_{cls}$ is used, the number of candidates for potential novel proposals is smaller, which is insufficient to optimize objectness in our framework. Hence, as expected, we observe a trade-off effect in this parametric choice, simular to the choice of the threshold value in the original Faster R-CNN pipeline. As the result indicates, $Thre_{cls} = 0.75$ is a reasonable value for filtering the candidate proposals relatively well which is the value we used in our experiments.

\begin{table}[t]
	\centering
	\resizebox{0.75\textwidth}{!}{
	\begin{tabular}{l|l|l|l|l|l}
        \toprule
		$Thre_{cls}$ & 0.05 & 0.25 & 0.5 & 0.75 & 0.95\\
		\midrule
		nAP & 60.5 & 61.2 & 62.1 & \textbf{62.9} & 61.4\\
		\bottomrule
	\end{tabular}
 }
	\caption{Ablation studies on different hyperparameter settings. The effect of adjusting the $Thre_{cls}$ is demonstrated. The highest nAP has been bolded.}
	\label{tab:ablation2}
\end{table}

\subsubsection{The Strategy for Ranking the Proposals}

As we explained, we select the top 1,000 proposals that are most likely to contain objects  in our training stage according to their objectness logit scores. Note that our proposed hierarchical ternary region proposal network (HTRPN) has three objectness scores: (${tobj_{pre}}^0$ which indicates the predicted score of being a non-object, ${tobj_{pre}}^1$ which represents the score of being a true object, and ${tobj_{pre}}^2$ which represents the score of being potential objects from unseen novel classes. We rank the proposals by combing their ${tobj_{pre}}^1$ and ${tobj_{pre}}^2$, marked as ${tobj_{pre}}^1 \oplus {tobj_{pre}}^2$. Operator $\oplus$ could be either arithmetical addition (${tobj_{pre}}^1 + {tobj_{pre}}^2$) or maximum ($max({tobj_{pre}}^1, {tobj_{pre}}^2)$).

\begin{figure}[t]
    \centering
    \includegraphics[width=80mm]{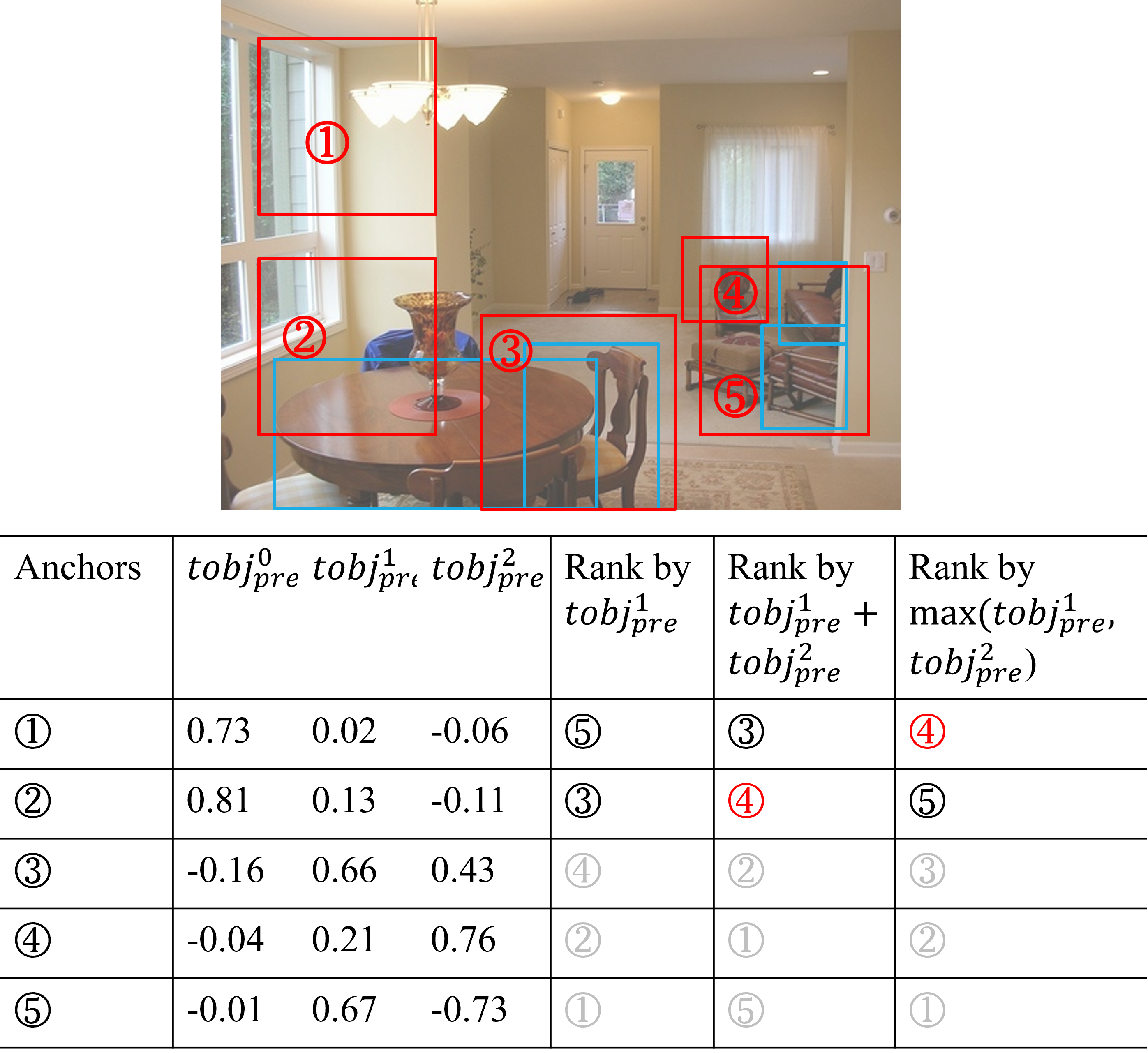}
    \caption{The 4 blue boxes are the ground truth boxes and the 5 red boxes are the proposals. The proposal \ding{175} contains an unlabeled novel object chair. The top 2 proposals that are ranked by ${tobj_{pre}}^1 \oplus {tobj_{pre}}^2$ could successfully include proposal \ding{175}.}
    \label{fig:maximum}
\end{figure}

The examples in Figure.~\ref{fig:maximum} demonstrate the difference between these two choices for the operator $\oplus$. In the image, labeled base class objects (tables and chairs) are in blue boxes. For five exemplary proposals \ding{172} to \ding{176} in red boxes with their ternary objectness logit scores, we intend to pick the top 2 of them. If we rank the proposals only by ${tobj_{pre}}^1$, then the proposals that contain unlabeled potential objects (e.g. proposal \ding{175}) will more likely be ignored since they are trained to have higher ${tobj_{pre}}^2$ score instead of ${tobj_{pre}}^1$. Therefore, it is necessary to take the ${tobj_{pre}}^2$ score into account, and according to our HTRPN, the objectness of an object should be presented by ${tobj_{pre}}^1$ and ${tobj_{pre}}^2$ together so that the rank of the potential proposal such as \ding{175} could be significantly promoted. However, by the method of ${tobj_{pre}}^1 + {tobj_{pre}}^2$, the rank of some proposals will be negatively influenced by the negative logit values, such as the proposal \ding{176}. On the contrary, by applying the maximum between ${tobj_{pre}}^1$ and ${tobj_{pre}}^2$, the rank of proposal \ding{176} will not be degraded by the singular negative values.

\subsubsection{Extra Computation Overhead}

FSCE~\cite{Sun21} has the closest architecture to our  proposed method. Our architecture is larger which may raise questions about its inference efficiency. For this reason, we have reported various performance metrics for our method in Table.~\ref{tab:computation overhead}, including the number of parameters, model size, and inference speed, and compared them with FSCE. By conducting these comparisons, we aim to assess the effectiveness of our method in terms of its ability toaccurately and efficiently encode features for FSOD tasks. The reported values are based on the pre-training of the PASCAL VOC dataset and a 5-shot fine-tuning scenarios. We observe that our method primarily takes effect during the training process without significantly affecting the inference stage, which ensures that our method does not slow down the inference speed .Our algorithm also minimizes any potential interference between the training and inference stages. By focusing on these two distinct phases of the process, we are able to deliver an   efficient and effective solution for the FSOD tasks compared to FSCE despite using a larger model.

\begin{table}[hb!]
	\centering
	\resizebox{\linewidth}{!}{
		\begin{tabular}{l|ll|ll|ll}
			\toprule
			\multirow{2}{*}{Method} & \multicolumn{2}{c|}{\makecell{Trainable number of\\ parameters}} & \multicolumn{2}{c|}{\makecell{Model size \\ (MB)}} & \multicolumn{2}{c}{\makecell{Inferencing speed \\ (s/img)}}\\
             & Pre-train & Fine-tune & Pre-train & Fine-tune & Pre-train & Fine-tune\\
			\midrule
            FSCE & 60,084,315 & 19,118,303 & 482.6 & 323.5 & 0.170 & 0.171 \\
			\midrule
            Ours & 76,343,137 & 34,196,325 & 612.6 & 430.7 & 0.170 & 0.170 \\
			\bottomrule
		\end{tabular}
    }
	\caption{\small Computation overhead between FSCE vs our method.}
	\label{tab:computation overhead}
\end{table}

\subsubsection{Recall of Novel Instances}

To assess and demonstrate the effectiveness of our proposed HTRPN, we have conducted a recall evaluation of novel objects in our method compared to   FSCE  ~\cite{Sun21}.  The experiment was conducted within the context of the 5-shot setting of the PASCAL VOC dataset, specifically using the split1 configuration. We also set the average recall value of novel object ($nAR$) on PASCAL VOC testing set to $IoU = 0.5$. The results are  reported in Table.~\ref{tab:recall}. We observe that our proposed method is effective in improving the recall of novel objects and offers an advantage compared to FSCE.  This experiment reinforces the effectiveness of our HTRPN in addressing the challenge of accurately identifying and recalling novel objects within the visual context, thereby contributing to improved FSOD performance.

\begin{table} 
	\centering
 \resizebox{0.3\textwidth}{!}{
		\begin{tabular}{l|l|l}
			\toprule
			Method & FSCE & Ours\\
			\midrule
            nAR & 72.7 & 73.8 \\
			\bottomrule
		\end{tabular}}
	\caption{Recall of novel objects for FSCE vs our method.}
	\label{tab:recall}
\end{table}

\subsubsection{Magnitude of Potential Novel Objects}
\begin{figure}[t]
    \centering
    \begin{subfigure}{\linewidth}
        \centering
    	\includegraphics[width=100mm]{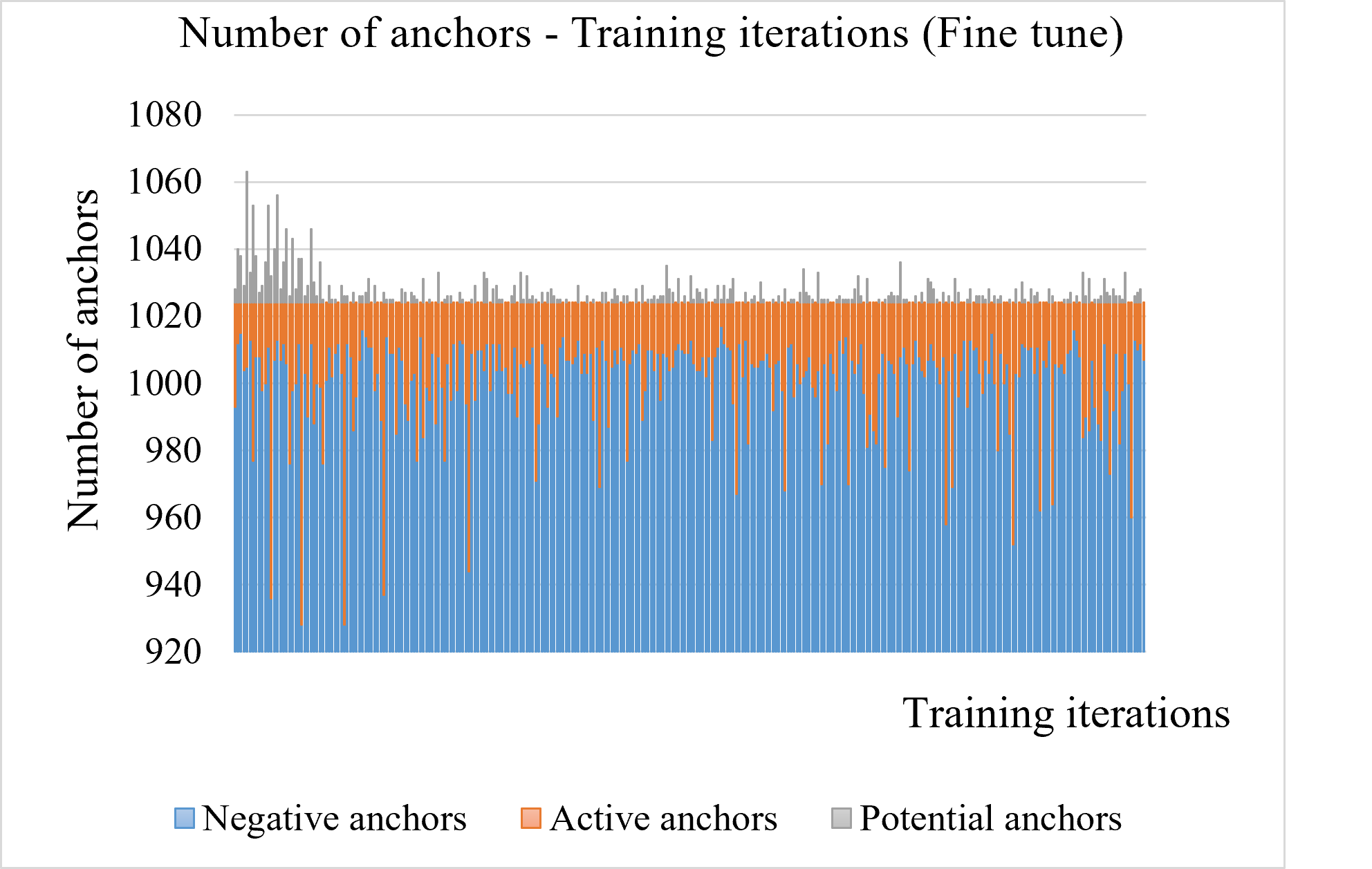}
    	\caption{Number of anchors - Training iterations chart of the pre-training stage.}
    	\label{fig:number of anchors pre train}
    \end{subfigure}
	\centering
    \begin{subfigure}{\linewidth}
        \centering
    	\includegraphics[width=100mm]{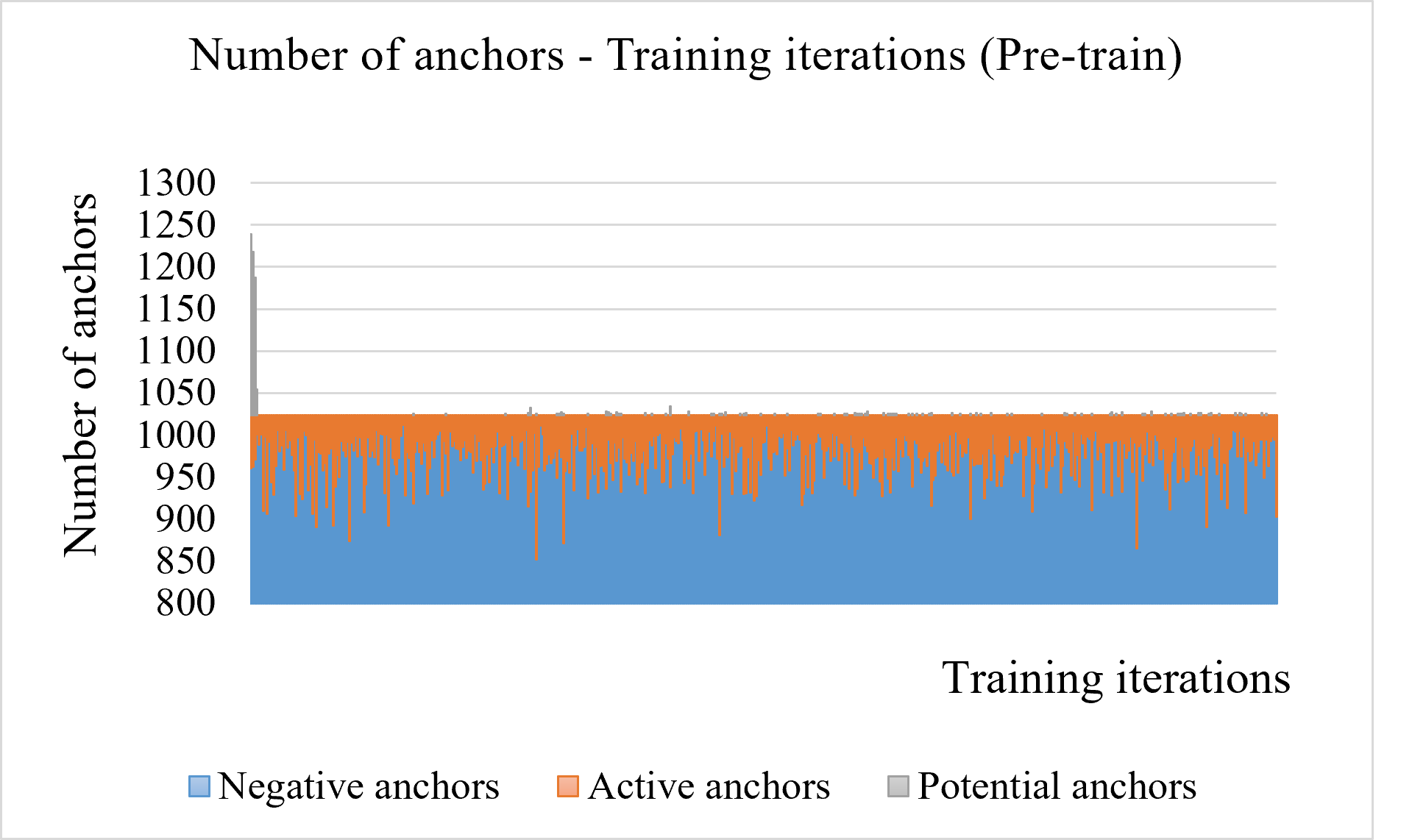}
    	\caption{Number of anchors - Training iterations chart of the fine-tuning stage.}
    	\label{fig:number of anchors fine tune}
    \end{subfigure}
	\caption{Magnitude of potential novel objects: the horizontal axis is the training iterations during fine-tuning, and the vertical axis indicates the number of anchors of each training iteration.}
	\label{fig:number of anchors}
\end{figure}

The core motivation of our work is the interference that instances from novel unseen can cause for FSOD. To demonstrate that we address this challenge by our training procedure,
 we record the number of potential objects during the pre-training and fine-tuning stages to demonstrate the statistical importance of the potential novel objects and by extension demonstrate that the additional complexities in our architecture address a practically noticeable challenge. We conduct this  experiments using the PASCAL VOC dataset in the 5-shot setting. The instance-level ternary classification threshold is fixed as $Thre_{cls}=0.75$ . The relation between the number of anchors and the training iteration in the pre-training stage is presented in Figure.~\ref{fig:number of anchors pre train}. In this experiment, the overall training anchors for each image is 256. Thus, for a batch size of 16 on 4 GPUs, the overall training anchors on a single GPU would be around $(16/4)*256=1024$. The negative anchors (blue bars) form the majority, and the number of active anchors (orange bars) are around 0 to 50 for each iteration, which translates into about 4\% of all anchors. However, the number of potential anchors (gray bars) converges with training iterations and stays at around 0 to 5 for each iteration. This observation indicates our model is getting more stable for recognizing the potential novel objects as more training iterations are performed. Furthermore, as shown in Figure.~\ref{fig:number of anchors fine tune}, during 5-shot fine-tuning, the trend of the potential anchors is similar to the pre-training stage. However, when transferring to novel classes, the pre-trained model tends to predict some labeled seen novel objects as potential unseen novel objects at the beginning of the fine-tuning stage, which could be considered as the inertia of the pre-trained model. Although the number of potential anchors is relatively high in the first several iterations of training, our training procedure mitigates this problem. This observation also supports our hypothesis that all true/potential positive ternary objectness of anchors should be presented by the combination of ${tobj_{pre}}^1$ and ${tobj_{pre}}^2$. We also observe that the number of potential anchors is gradually stabilized to around 0 to 15 for each iteration, which is about 1.5\% of all anchors in an iteration.

\begin{figure}[t]
    \centering
    \includegraphics[width=120mm]{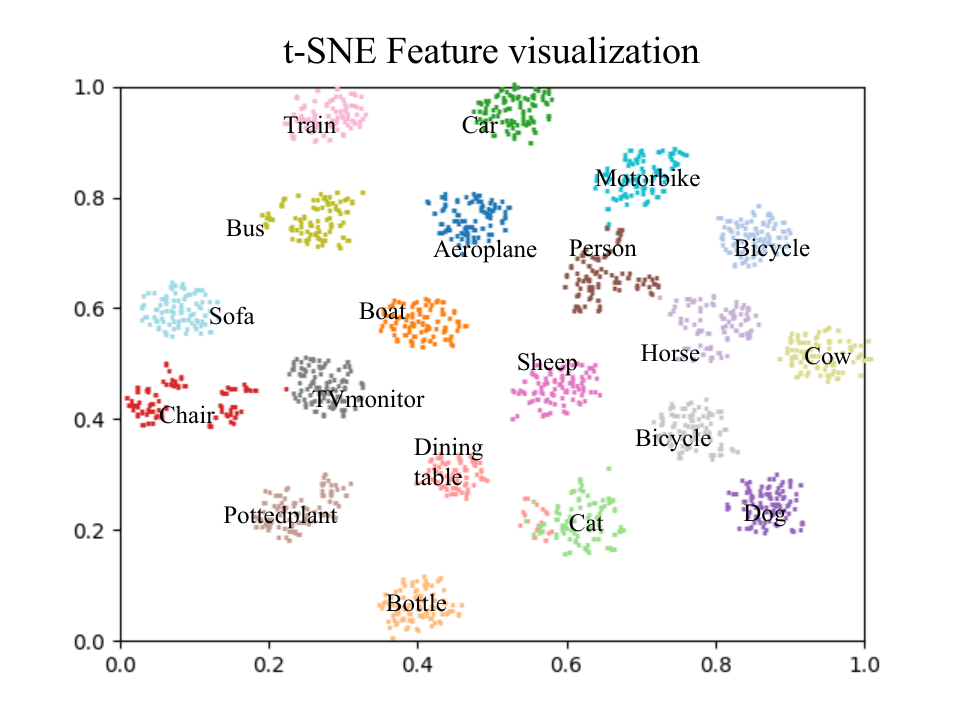}
    \caption{The t-SNE visualization of our HTRPN.}
    \label{fig:tsne}
\end{figure}

\subsubsection{Feature Representation Analysis}

Successful FSOD would be feasible in our archtecture, if the extracted features by  HTRPN describe the classes well, i.e., representations for data points that belong two the same class become separated from the rest of classes. To study this aspect, Figure.~\ref{fig:tsne} shows the t-SNE visualization of clustering of our model for the split1 5-shot scenario of the PASCAL VOC dataset. In this figure, we can observe the clusters formed by the algorithm and the corresponding color for each data point indicates its label. We can see that separated clusters are formed, where each one represents a different class in the dataset. A denser cluster indicates smaller intra-class distance and better recognition ability of its category because larger margins make the model robust with respect to domain shift. 
The t-SNE visualization indicates that our model exhibits a commendable ability to cluster distinct categories, demonstrating its effectiveness in identifying and separating them for FSOD.

\section{Conclusions}
We identified and addressed an unexplored challenge for FSOD.
 Our proposed architecture enhances the quality of FSOD model by improving the R-CNN-based architecture. Our approach focused on addressing the challenge of objectness inconsistency due to potential unlabeled novel objects that belong to unseen classes. By studying this phenomenon, we were able to improve the accuracy and robustness of the base FSOD model. We implemented a balance anchor sampling strategy to improve the ability of the model to identify anchors that may contain objects from unseen classes. In addition, we proposed HTRPN which leads to the recognition ability of potential novel objects and further enhances the objectness consistency. As our results demonstrate, this approach can mitigate model confusion about the novel classes and lead to noticeable improvements in the  object detection performance, and we are excited. Future research includes    extensions of our idea to identify novel unseen classes in a zero-shot learning setting to address a similar challenge.


\bibliography{ref}

\end{document}